\documentclass{mva_style}
\usepackage{graphicx}
\usepackage{amsmath}
\usepackage{float}
\restylefloat{figure}
\usepackage{color}
\usepackage{booktabs}
\usepackage{tabularx}
\usepackage{multirow}
\usepackage{moresize}
\usepackage{caption}
\usepackage{array}
\usepackage{afterpage}
\newcommand{\addpic}[1]{\includegraphics[width=4em]{#1}}
\newcommand{\verns}[1]{\rotatebox{90}{\tiny{#1}}}

\newcommand{\ver}[1]{\rotatebox{90}{\tiny{\hspace{0.2 cm}#1}}\hspace{0.2cm}}

\newcolumntype{Y}{>{\centering\arraybackslash}X}
\newcommand\notsotiny{\@setfontsize\notsotiny\@vipt\@viipt}

\finalcopy %Uncomment this line for the Camera-Ready Manuscript

\begin{document}
\title{Colored Transparent Object Matting from a Single Image Using Deep Learning}

%\author{
%  Jamal Ahmed Rahim\\
%  University of Hong Kong\\
%  Hong Kong\\
%  {\tt jamal@connect.hku.hk}\\
%  \and
%  Kwan-Yee K. Wong\\
%  University of Hong Kong\\
%  Hong Kong\\
%  {\tt kykwong@cs.hku.hk}\\
%}

\author{
  Jamal Ahmed Rahim, Kwan-Yee K. Wong\\
  University of Hong Kong\\
  Hong Kong\\
  {\tt \{jarahim, kykwong\}@cs.hku.hk}\\
}

\maketitle

\section*{\centering Abstract}
\textit{
This paper proposes a deep learning based method for colored transparent object matting from a single image. Existing approaches for transparent object matting often require multiple images and long processing times, which greatly hinder their applications on real-world transparent objects. The recently proposed TOM-Net can produce a matte for a colorless transparent object from a single image in a single fast feed-forward pass. In this paper, we extend TOM-Net to handle colored transparent object by modeling the intrinsic color of a transparent object with a color filter. We formulate the problem of colored transparent object matting as simultaneously estimating an object mask, a color filter, and a refractive flow field from a single image, and present a deep learning framework for learning this task. We create a large-scale synthetic dataset for training our network. We also capture a real dataset for evaluation. Experiments on both synthetic and real datasets show promising results, which demonstrate the effectiveness of our method.}

\section{Introduction}
Image matting is an important problem in the field of image processing. It refers to the process of extracting a foreground matte from an image by identifying pixels of the foreground object and estimating their opacities. The extracted matte can be composited onto different backgrounds to produce new images based on the matting equation:
\begin{equation}
I_{i,j} = \alpha_{i,j}F_{i,j} + (1-\alpha_{i,j})B_{i,j}, \hspace{0.5 cm} \alpha_{i,j} \in [0, 1]
\end{equation}
where $I_{i,j}$, $\alpha_{i,j}$, $F_{i,j}$  and  $B_{i,j}$ denote the composite color,  opacity, foreground color, and background color at pixel $(i,j)$ respectively.

Image matting is widely used in image and video editing. However, existing methods are often designed for opaque objects and cannot handle transparent objects. Note that the appearance of a transparent object depends on how it refracts light from the background. Obviously, estimating object opacity alone is not sufficient to model the appearance of a transparent object. There exist a number of methods for transparent object matting \cite{trmat1, trmat2, trmat3, trmat4}. They often require a huge number of images, specially designed background images, and a long processing time to extract an environment matte for the transparent object.

Recently, with the great success of deep learning in both low-level and high-level vision tasks, deep learning framework has been introduced to transparent object matting. Chen~\textit{et al.}~\cite{TOM-Net} introduced a two-stage framework for learning transparent object matting from a single image. However, their method assumes colorless transparent objects and cannot handle colored transparent objects. This limits its applications on real-world objects.

In this paper, we address the problem of colored transparent object matting from a single image. We extend TOM-Net to handle colored transparent object by modeling the intrinsic color of a transparent object with a color filter. We formulate the problem of colored transparent object matting as simultaneously estimating an object mask, a color filter, and a refractive flow field from a single image, and present a two-stage deep learning framework, named {CTOM-Net}, for learning this task. We create a large-scale synthetic dataset for training our network. We also capture a real dataset for evaluation. Promising results have been achieved on both synthetic and real datasets, which demonstrate the effectiveness of our method.

\section{Image Formation Model}
In ~\cite{TOM-Net}, Chen~\textit{et al.} formulated colorless transparent object matting as
\begin{equation}
I_{i,j} = (1 - m_{i,j})B_{i,j} + m_{i,j} \cdot \rho_{i,j} \cdot M(T,R_{i,j}), \label{eq:colorless_matting}
\end{equation}
where $I_{i,j}$,  $m_{i,j} \in \{0, 1\}$, $B_{i,j}$ and $\rho_{i,j} \in [0, 1]$ denote the composite color, foreground ($m_{i,j}=1$) / background ($m_{i,j}=0$) membership, background color and light attenuation index at pixel $(i,j)$ respectively. $M(T,R_{i,j})$ is a bilinear sampling operation at location $(i+R_{i,j}^x, j+R_{i,j}^y)$ on the background image $T$. The matte can be estimated by solving $m_{i,j}$, $\rho_{i,j}$ and $R_{i,j}$  for each pixel in the input image, and these result in an object mask, an attenuation map and a refractive flow field respectively.

In order to take color into account, we need to introduce color to (\ref{eq:colorless_matting}). If we model the object with a constant color, we will fail to capture color variations in the object and lose some valuable information. We therefore propose to estimate a color value for each pixel and cast this problem as estimating a color filter. In order to keep the solution space small, we integrate the color filter with the attenuation mask, and formulate colored transparent object matting as
\begin{equation}
I_{i,j} = (1-m_{i,j})B_{i,j} + m_{i,j} \cdot min(C_{i,j}, M(T,R_{i,j})),
\end{equation}
where $C_{i,j}$ denotes the``color filter' at pixel $(i,j)$. The matte can now be estimated by solving $m_{i,j}$, $C_{i,j}$ and $R_{i,j}$ for each pixel in the input image.

\begin{figure*}[h!]
 \includegraphics[width=\textwidth]{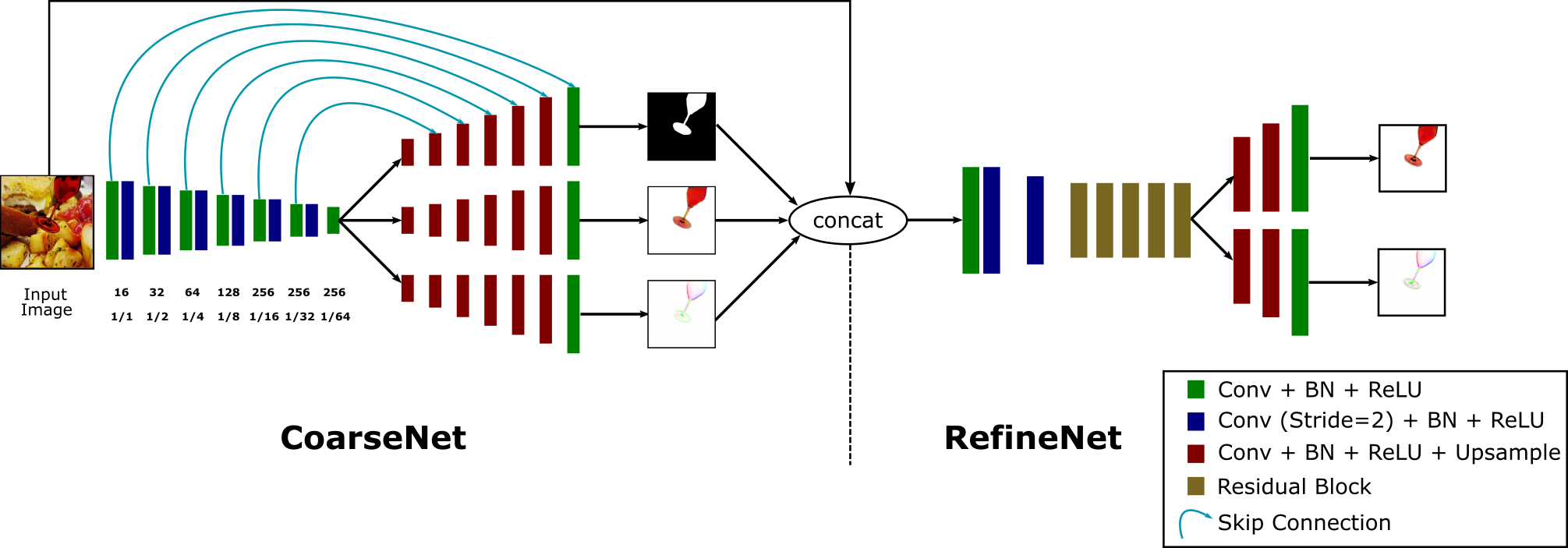}
 \caption{Our two-stage framework. The first value under an encoder block is the number of features extracted and the second value is the resolution relative to the input image. All filters are of size $3\times3$.}
  \label{fig:framework}
\end{figure*}

\section{Our Regression Framework}
Based on the TOM-Net \cite{TOM-Net} architecture, we propose a framework called CTOM-Net (see Fig.~\ref{fig:framework}) for learning colored transparent object matting. Like TOM-Net, CTOM-Net is a two-stage framework composed of the CoarseNet (1st stage) and the RefineNet (2nd stage). CoarseNet is an encoder-decoder multi-scale network. It takes a single image as input and outputs an object mask, a color filter, and a refractive flow field. It can produce a robust object mask, but the color filter and refractive flow field lack local structural details. This may result in noisy composites at test time. RefineNet, which is a residual network \cite{refnet}, is introduced to refine the color filter and refractive flow field estimated by CoarseNet. It takes the input image and the outputs of the CoarseNet as input, and outputs a sharper color filter and a more detailed refractive flow field.

\subsection{CoarseNet}
Inspired by the results of TOM-Net and U-Nets \cite{mirror-link, docunet} in problems requiring dense predictions, we adopt a mirror-link CNN \cite{mirror-link} for CoarseNet. Specifically, CoarseNet has one shared encoder and three parallel decoders for regressing an object mask, a color filter and a refractive flow field respectively (see Fig.~\ref{fig:framework}).

\subsubsection{Loss Function}
The learning of CoarseNet is supervised by the ground-truth matte using an object mask segmentation loss $L_{m}$, a color filter regression loss $L_{c}$, and a refractive flow field regression loss $L_{r}$ respectively. Besides, we also include an image reconstruction loss $L_{i}$ to ensure the reconstructed image\footnote{image generated by compositing an estimated matte on the same background as the input} being as close to the input as possible. The overall loss function for CoarseNet is given by
\begin{equation}
	L = \alpha_mL_m + \alpha_cL_c + \alpha_rL_r + \alpha_iL_i,
\end{equation}
where $\alpha_m$, $\alpha_c$, $\alpha_r$, $\alpha_i$ are the weights for the corresponding loss terms respectively.\\

\noindent{\bf Object Mask Segmentation Loss}
The object mask has a dimension of $2 \times W \times H$, with $W$ and $H$ being the width and height of the input image respectively. The problem of estimating an object mask is modeled as a binary classification problem, with the output being normalized by the \textit{softmax} function. We compute the object mask segmentation loss using the binary cross-entropy function as
\begin{multline}
L_m =\\ - \frac {1}{HW} \sum_{i,j} (\tilde{M}_{i,j}\log(P_{i,j}) + (1 - \tilde{M}_{i,j})\log(1-P_{i,j})),
\end{multline}
where $\tilde{M}_{i,j} \in \{0, 1\}$ denotes the ground-truth binary mask at pixel $(i, j)$, and $P_{i,j} \in [0, 1]$ denotes the predicted probability of being foreground at pixel $(i, j)$ .\\

\noindent{\bf Color Filter Regression Loss}
The color filter is a three-channel mask with a dimension of $3 \times W \times H$. Its accuracy can be measured by its similarity with the ground-truth color filter. We compute the color filter regression loss using the mean squared error function as
\begin{equation}
L_c =  \frac {1}{HW} \sum_{i, j}||\tilde{C}_{i, j} - C_{i, j}||_{2}^{2},
\end{equation}
where $\tilde{C}_{i, j}$ and $C_{i, j}$ are the ground-truth and predicted color filter at pixel $(i,j)$ respectively.\\

\noindent{\bf Refractive Flow Field Regression Loss}
The refractive flow field has a dimension of $2 \times W \times H$, with one channel for horizontal displacements, and another channel for vertical displacements. To restrict the horizontal and vertical displacements to the ranges $[-W, W]$ and $[-H, H]$ respectively, we normalize the output by the \textit{tanh} function and multiply the result by the width and height of the input image accordingly. We compute the refractive flow field regression loss using the average end-point error (EPE) as
\begin{equation}
L_r = \frac {1}{HW} \sum_{i,j} \sqrt{(\tilde{R}_{i, j}^{x} - R_{i, j}^{x})^{2} + (\tilde{R}_{i, j}^{y} - R_{i, j}^{y})^{2}},
\end{equation}
where $(\tilde{R}_{i, j}^{x}, \tilde{R}_{i, j}^{y})$ and $(R_{i, j}^{x}, R_{i, j}^{y})$ are the ground-truth and predicted displacements at pixel $(i,j)$ respectively.\\

\noindent{\bf Image Reconstruction Loss}
The composite image has a dimension same as the input image (i.e., $3 \times W \times H$). We compute the image reconstruction loss using the mean squared error function as
\begin{equation}
L_i =  \frac {1}{HW} \sum_{i, j}||\tilde{I}_{i, j} - I_{i, j}||_{2}^{2},
\end{equation}
where $\tilde{I}_{i, j}$ and $I_{i, j}$ are the input image and the reconstructed image at pixel $(i,j)$ respectively.\\

\noindent{\bf Implementation Details}
We empirically set $\alpha_m$ to 0.25, $\alpha_c$ to 1.1, $\alpha_r$ to 0.01 and $\alpha_i$ to 1.0. It takes approximately 2 days to train our network on a single GPU (Nvidia GTX 1080 Ti).

\subsection{RefineNet}
Following the TOM-Net architecture, we adopt a residual network to refine our output. The input to RefineNet is the predicted matte from CoarseNet concatenated with the input image, and the outputs are a refined color filter and a refined refractive flow field. The object mask is not refined by the RefineNet as its prediction by the CoarseNet is already very good and robust.

\subsubsection{Loss Function}
The loss for RefineNet is based only on the color filter and refractive flow field predictions. The loss functions for color filter $L_c$ and refractive flow field $L_r$ remain the same as in CoarseNet. The overall loss function for the RefineNet is
\begin{equation}
L^r = \alpha_c^rL_c + \alpha_r^rL_r.
\end{equation}
In our implementation, we empirically set both  $\alpha_c^r$ and $\alpha_r^r$ to 1.0.

\section{Dataset}
\noindent{\bf Synthetic Dataset}
We perform supervised learning for colored transparent object matting, and therefore we need the ground-truth mattes of our inputs for training. Since it is very difficult and time-consuming to obtain ground-truth mattes for real images of transparent objects, we created a synthetic dataset using POV-Ray \cite{POV-Ray} with shapes taken from the publicly available dataset created in \cite{TOM-Net}, and background images from the Microsoft COCO dataset \cite{COCO}. The shapes were taken from three categories, namely glass, glass with water and lens. They were rendered in front of a random background image with random orientation, scale, position, and filter color. In order to make sure the network does not overfit the synthetic colored transparent objects, we also include a number of colorless transparent objects in the dataset. Table~\ref{dataset stats} shows the statistics of our synthetic dataset. The network trains very well on the synthetic dataset and generalizes well on both colorless and colored transparent objects.\\

\begin{table}[t]
	\scriptsize
	\centering
	\caption{Our Synthetic Dataset Statistics.}
        \begin{tabular}{c|ccc|ccc}
			\toprule
			\multirow{2}{*}[-0.7 em]{Split} &
			\multicolumn{3}{c|}{Colored} &
			 \multicolumn{3}{c}{Colorless} \\
        &Glass & G\&W & Lens &Glass & G\&W & Lens\\ \midrule
		Train & 52,000& 26,000 &20,000 & 10,000& 10,000& 10,000 \\
		Test & 400& 400& 400& 400& 400& 400\\
		\bottomrule

    \end{tabular}

        \label{dataset stats}
    \end{table}

\noindent{\bf Real Dataset} In order to test the transferability of our model on real images, we also captured a real dataset of colored transparent objects. This dataset contains colored transparent objects from three categories, namely glass, glass with water and lens. Each example in the dataset has an image of a transparent object in front of a background and an image of the background alone. The background images are used for direct comparison as well for the generation of reconstructed images during testing. We capture 60 images each for the glass and glass with water category, and 30 images for the lens category. For colorless transparent objects, we use the real dataset made publicly available by \cite{TOM-Net}.\\

\noindent{\bf Data Augmentation} Extensive data augmentation is carried out at runtime during training. Apart from standard augmentation techniques such as rotation, scaling, shearing, cropping, and color jitter, we also try to bridge the gap between synthetic and real data by smoothing the borders of the transparent objects in the synthetic images and adjusting the ground truth color filter accordingly. Since real complex objects usually appear to be  more colored towards regions of total internal reflection and less colored in other regions, they would either require more careful data generation and/or augmentation. We therefore omit complex objects in training and testing,  and leave it for future works.

\section{Experimental Results}
Our evaluation shows promising results in colored transparent objects matting from a single image. It is worth noting that if the background and the transparent object are of similar color, it would be difficult to tell the extent to which the background color is being filtered. This implies that there is a limit to which such color filtering can be determined. Our results on synthetic and real datasets show that we have managed to reach this limit.

\subsection{Results on Synthetic Data}
\begin{figure*}[h!]
	\renewcommand{\thefigure}{2}
	\centering
        \begin{tabularx}{\textwidth}{c@{\hspace{0.0cm}}c@{\hspace{0.3cm}}c@{\hspace{0.3cm}}c@{\hspace{0.3cm}}c@{\hspace{0.3cm}}c}
			\toprule
        Input & Background & Ref. Flow (GT / Est) & Mask (GT / Est) & Color Filter (GT / Est) & Reconstruction\\ \midrule
		\multirow{2}{*}[1.2 em]{\verns{Colored Glass}\hspace{0.2 cm}\addpic{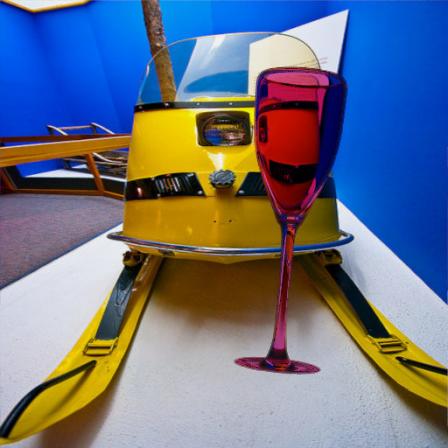}} & \multirow{2}{*}[1.0 em]{\addpic{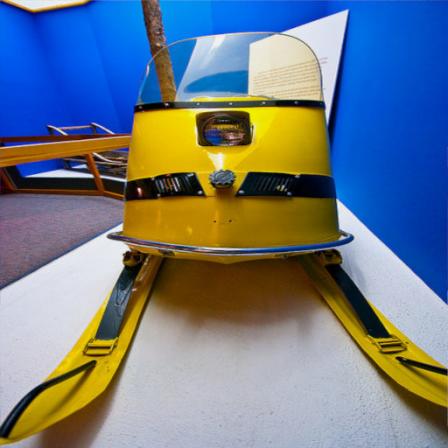}} & \ver{TOM-Net}\addpic{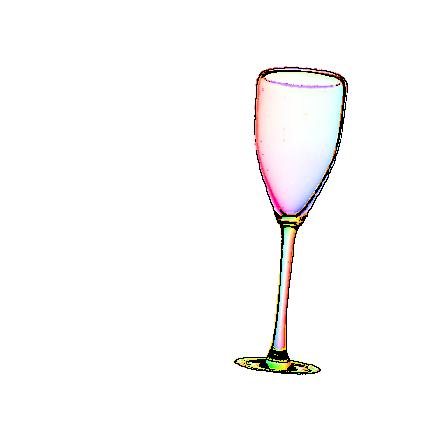} \addpic{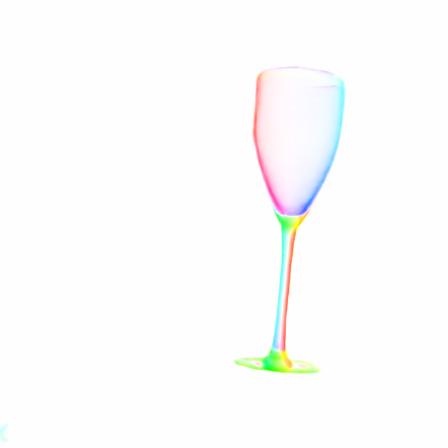} & \addpic{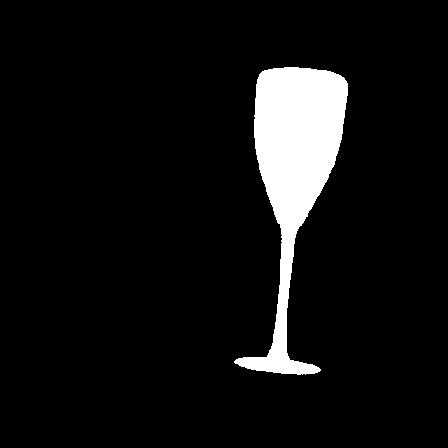} \addpic{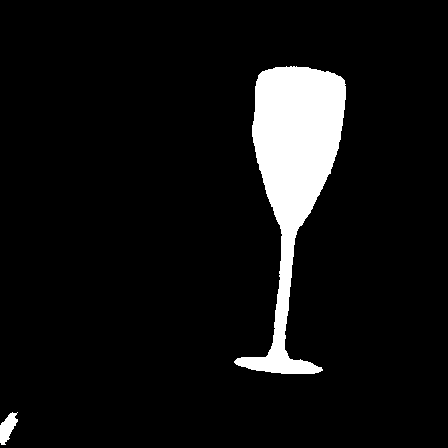} & \addpic{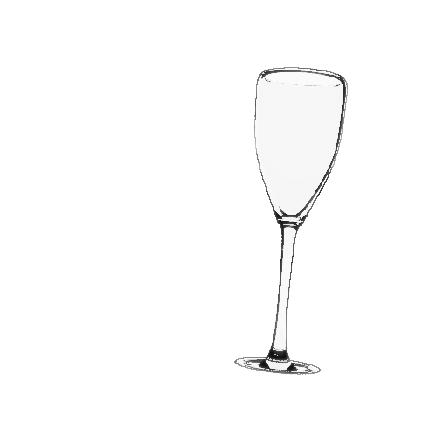} \addpic{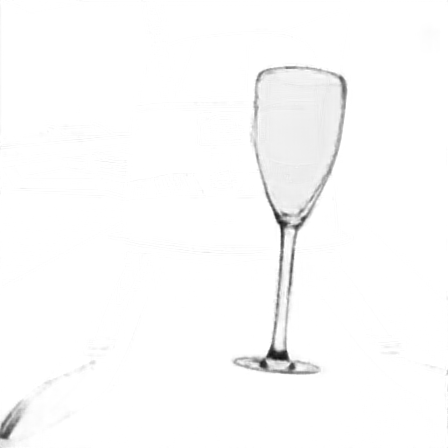} & \addpic{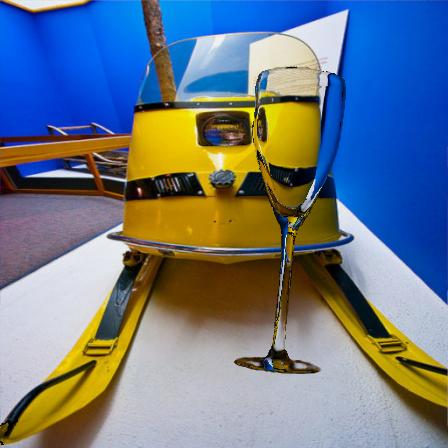} \\[-0.2 em]
  &  & \ver{CTOM-Net}\addpic{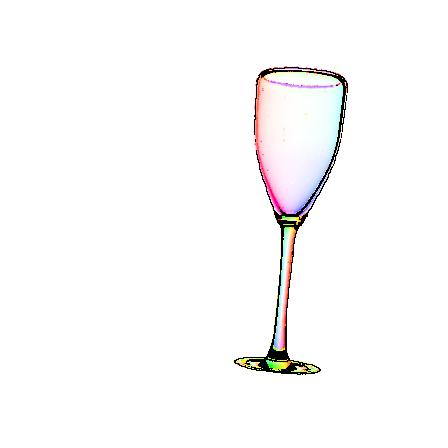} \addpic{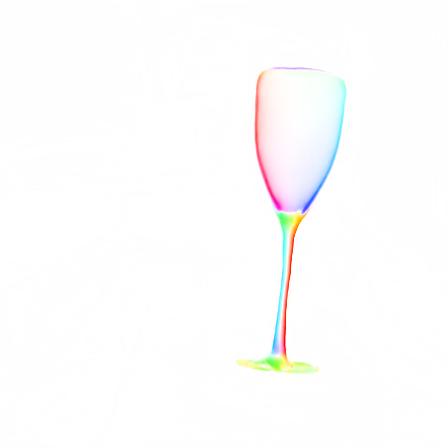} & \addpic{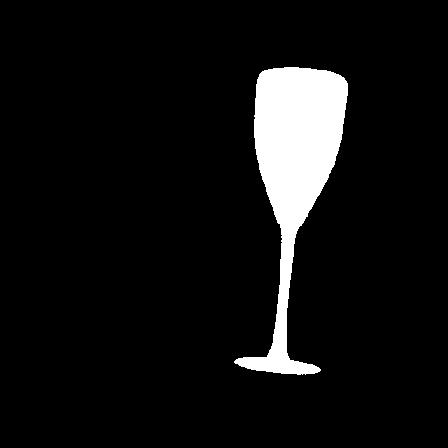} \addpic{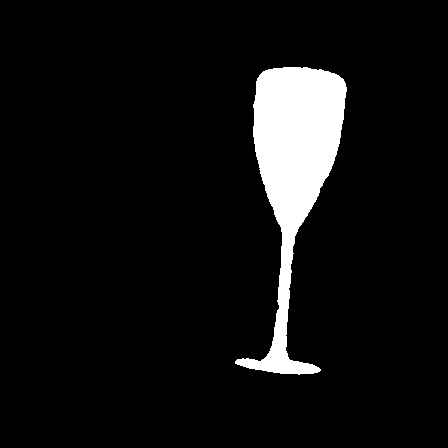} & \addpic{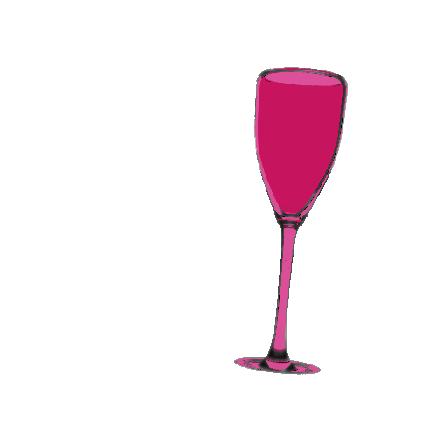} \addpic{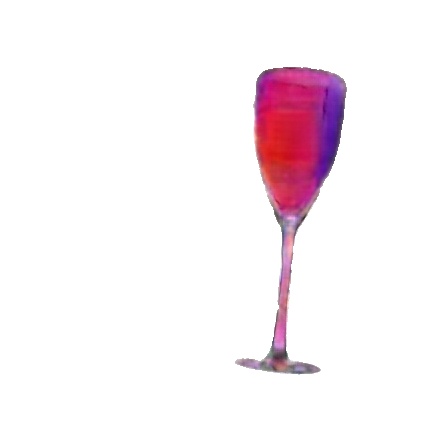} & \addpic{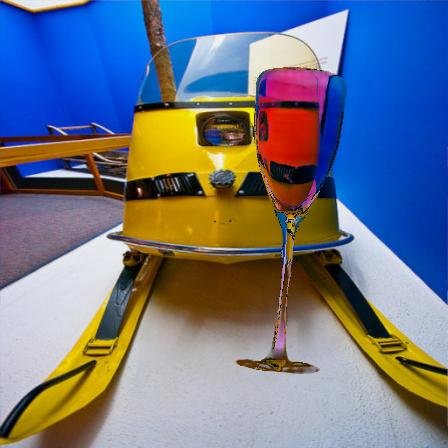} \\[-0.4 em] \midrule \\[-1.2 em]

  \multirow{2}{*}[1.5 em]{\verns{Colorless Glass}\hspace{0.2 cm}\addpic{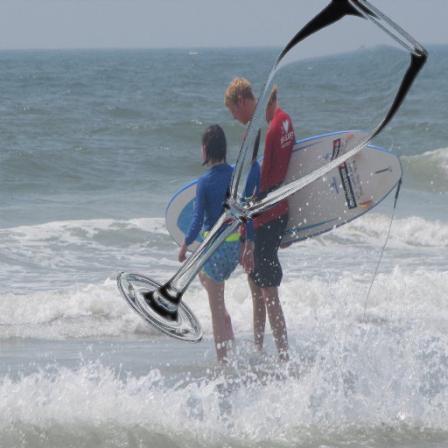}} & \multirow{2}{*}[1.0 em]{\addpic{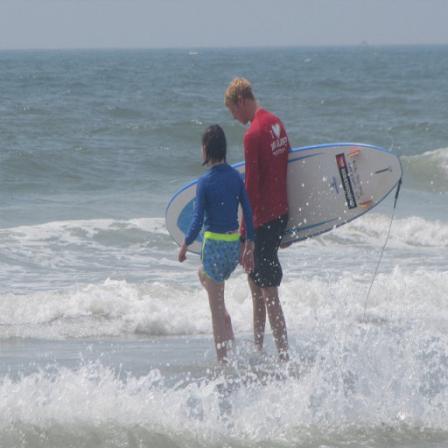}} & \ver{TOM-Net}\addpic{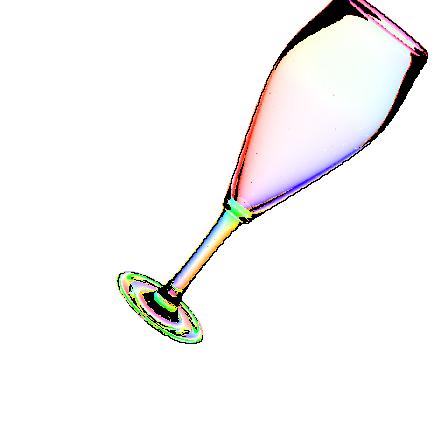} \addpic{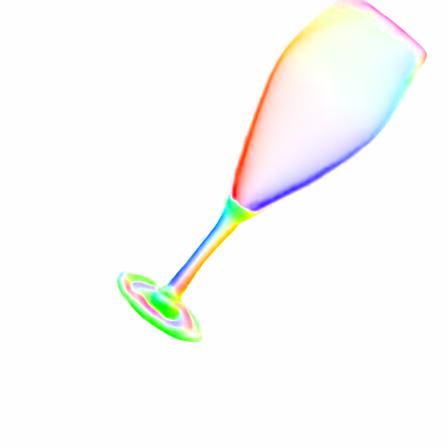} & \addpic{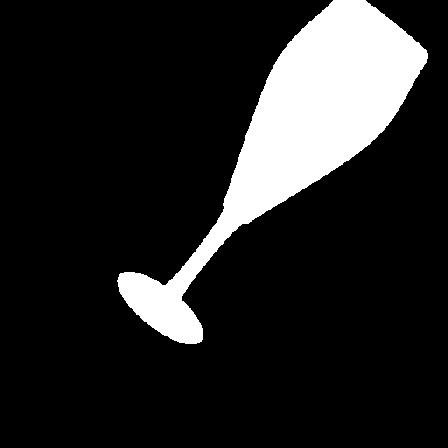} \addpic{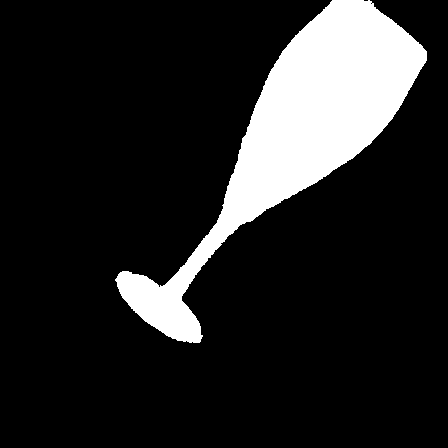} & \addpic{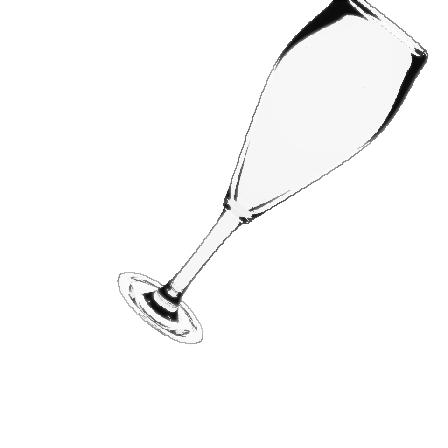} \addpic{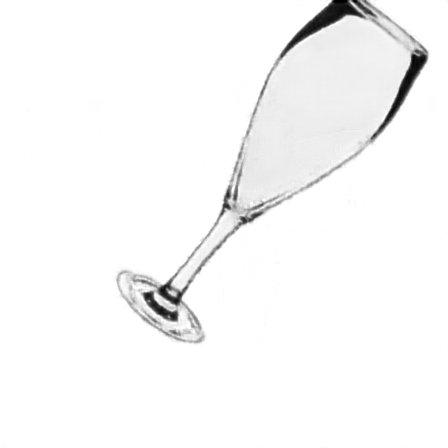} & \addpic{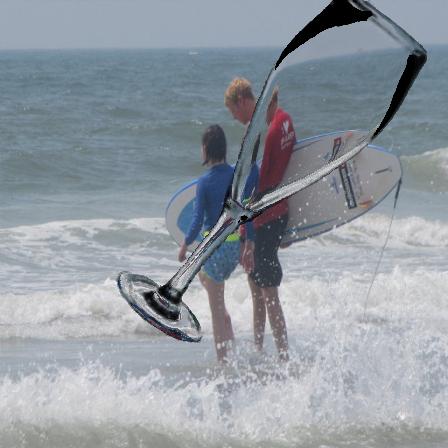} \\[-0.2 em]
&  & \ver{CTOM-Net}\addpic{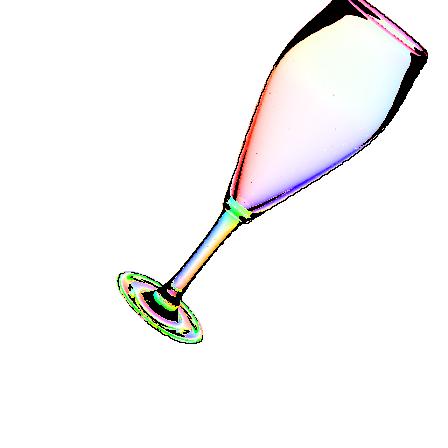} \addpic{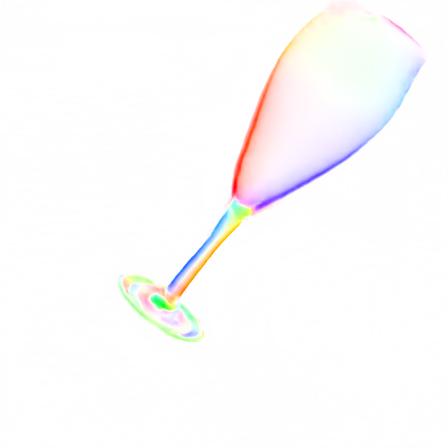} & \addpic{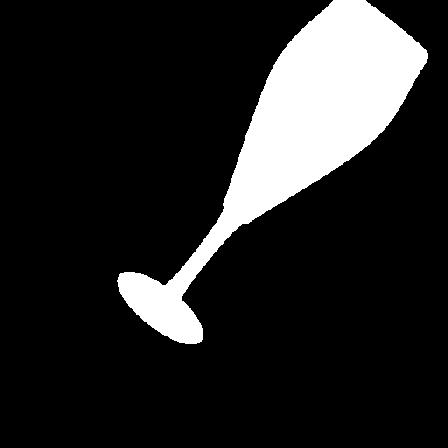} \addpic{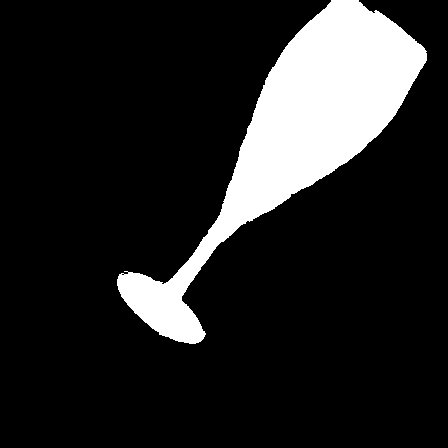} & \addpic{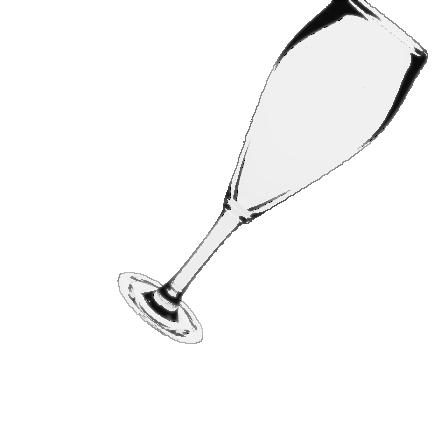} \addpic{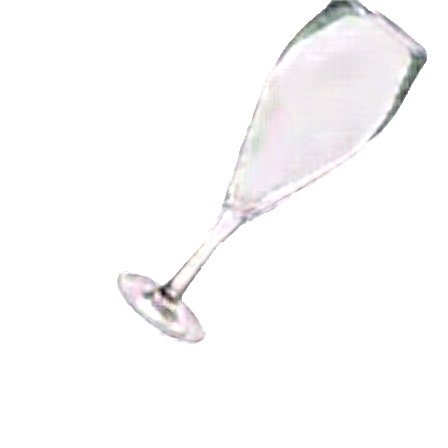} & \addpic{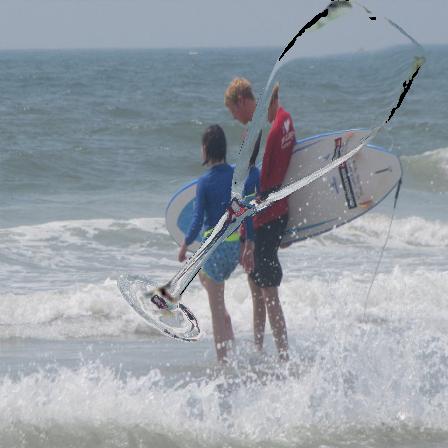}\\[-0.4 em] \midrule \\[-1.2 em]
\multirow{2}{*}[1.3 em]{\verns{Colored G\&W}\hspace{0.2 cm}\addpic{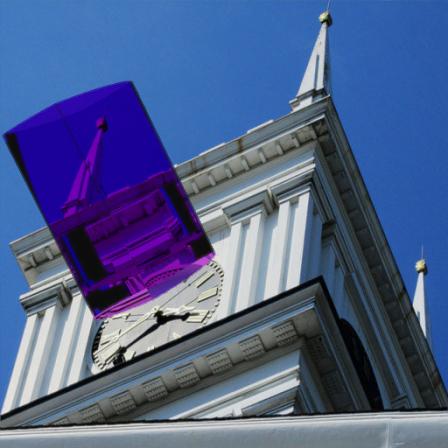}} & \multirow{2}{*}[1.0 em]{\addpic{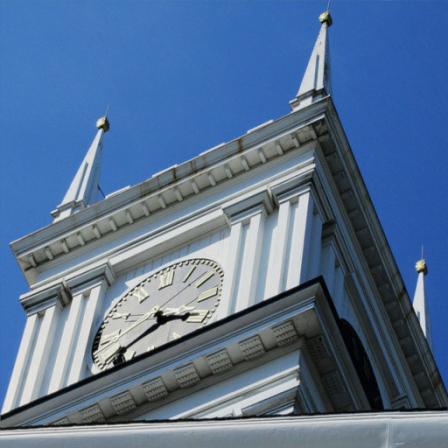}} & \ver{TOM-Net}\addpic{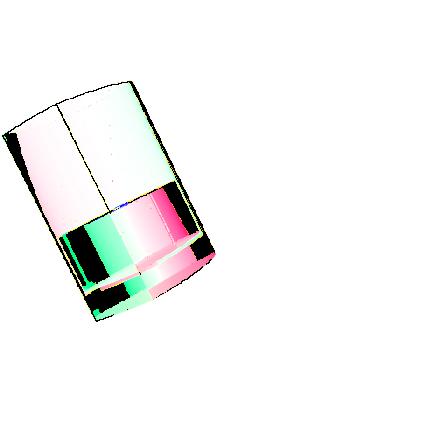} \addpic{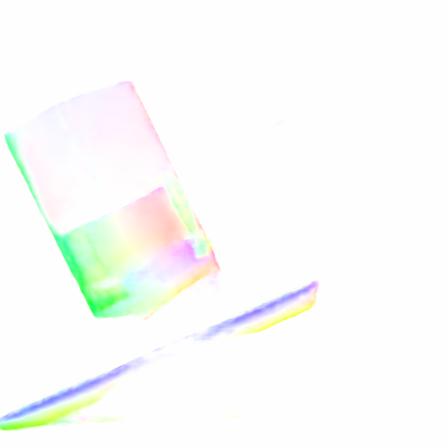} & \addpic{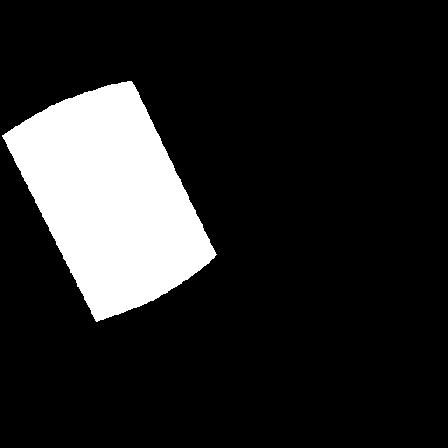} \addpic{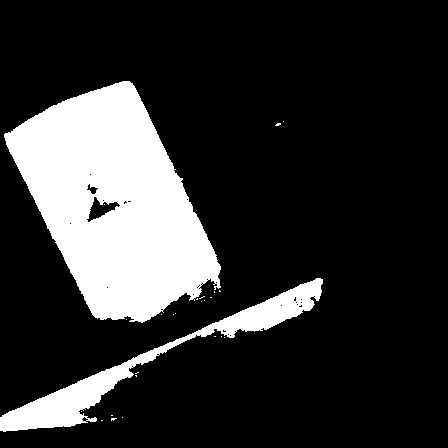} & \addpic{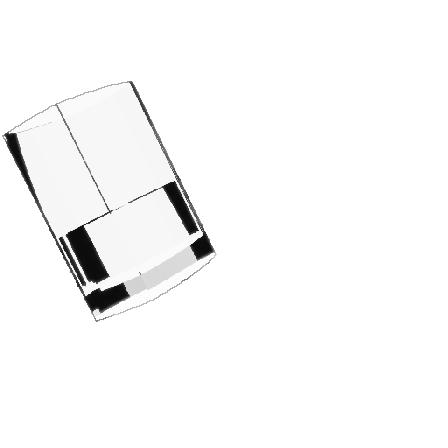} \addpic{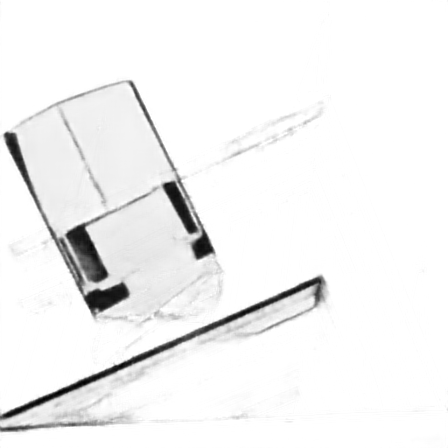} & \addpic{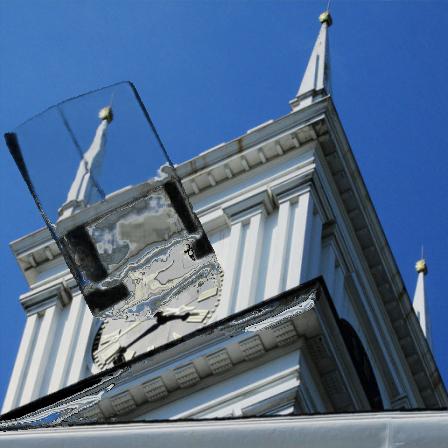} \\[-0.2 em]
&  & \ver{CTOM-Net}\addpic{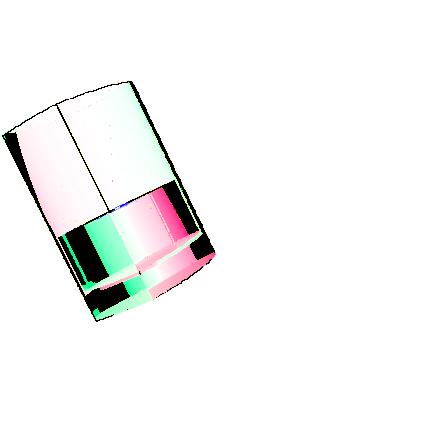} \addpic{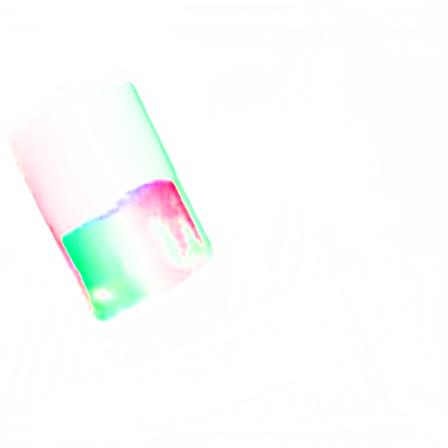} & \addpic{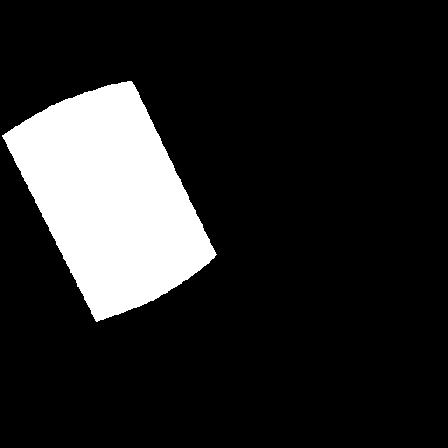} \addpic{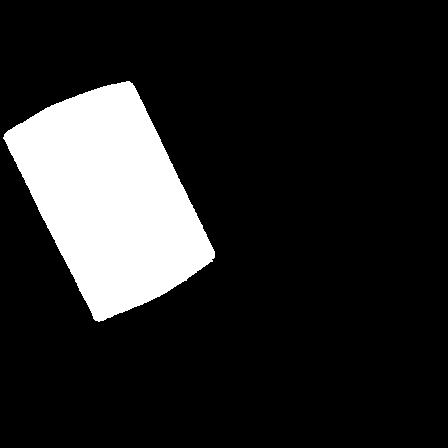} & \addpic{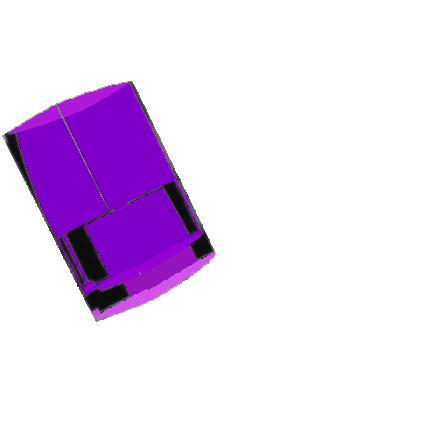} \addpic{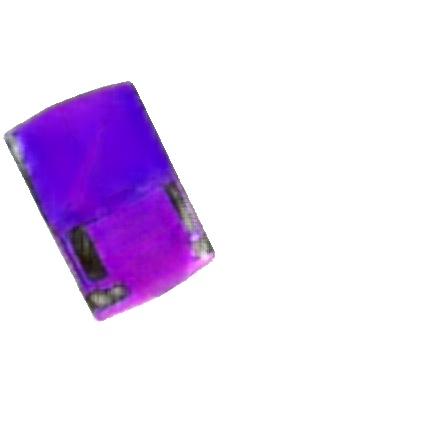} & \addpic{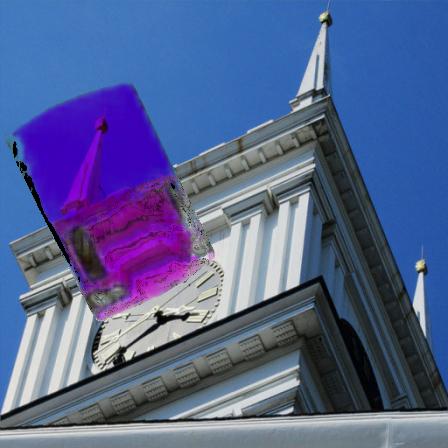}\\[-0.4 em] \midrule \\[-1.2 em]

\multirow{2}{*}[1.6 em]{\verns{Colorless G\&W}\hspace{0.2 cm}\addpic{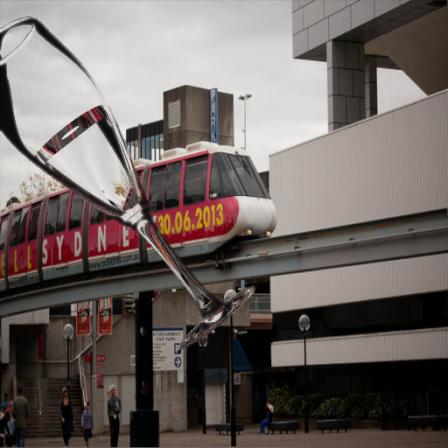}} & \multirow{2}{*}[1.0 em]{\addpic{}} & \ver{TOM-Net}\addpic{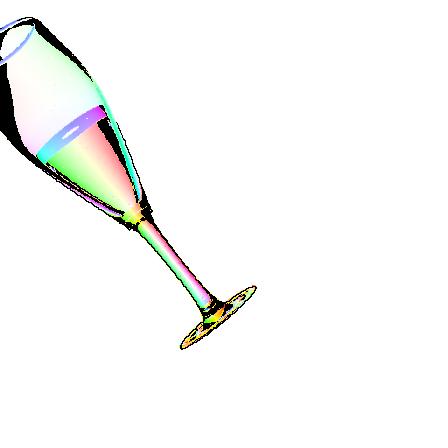} \addpic{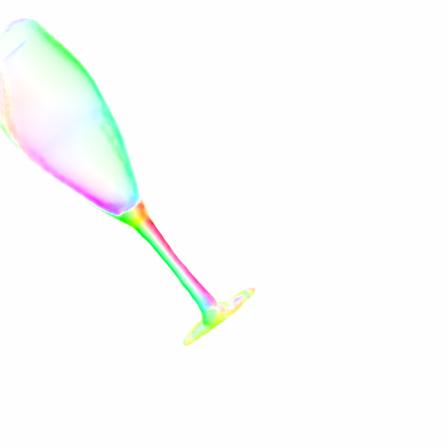} & \addpic{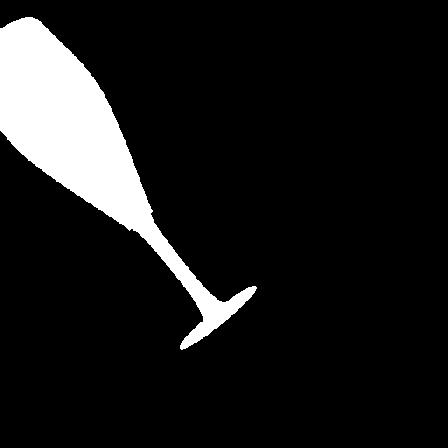} \addpic{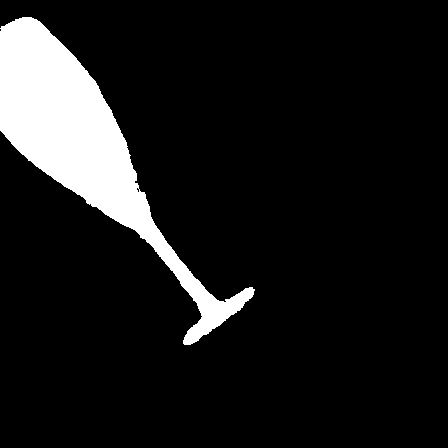} & \addpic{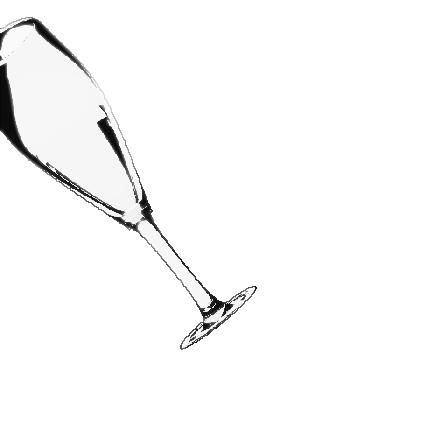} \addpic{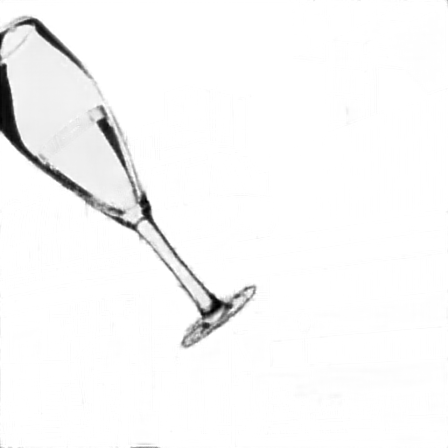}  & \addpic{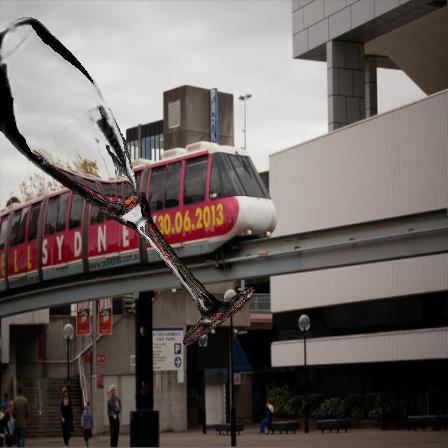} \\[-0.2 em]
&  & \ver{CTOM-Net}\addpic{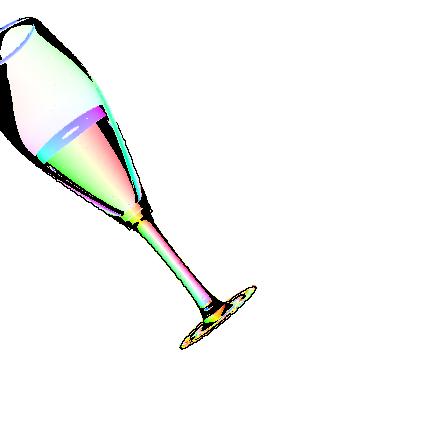} \addpic{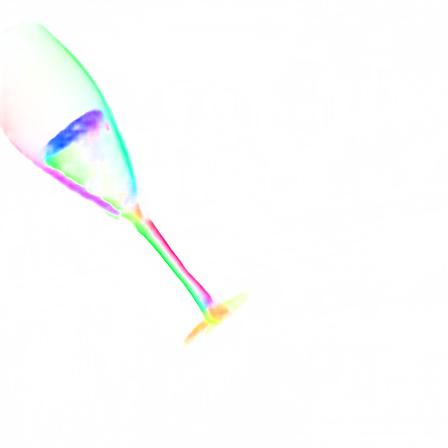} & \addpic{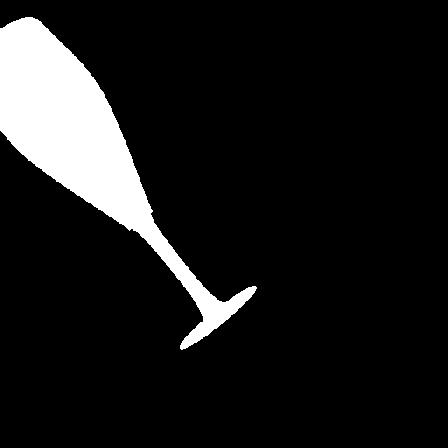} \addpic{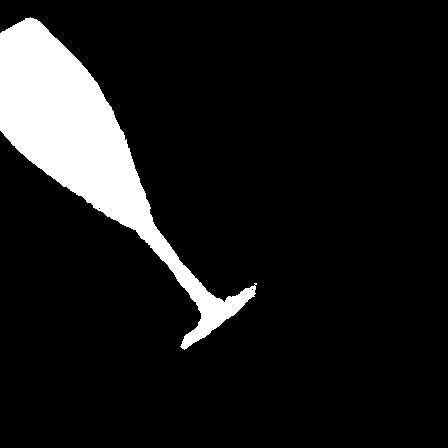} & \addpic{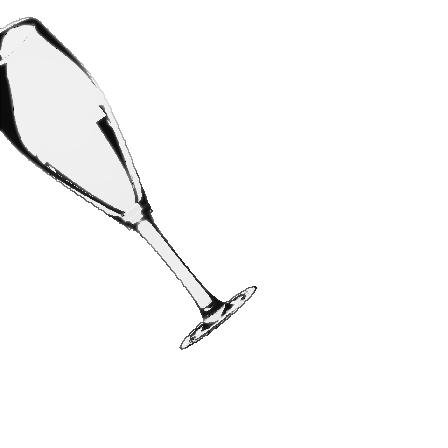} \addpic{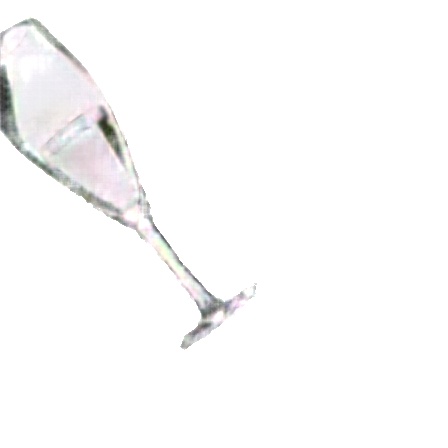} & \addpic{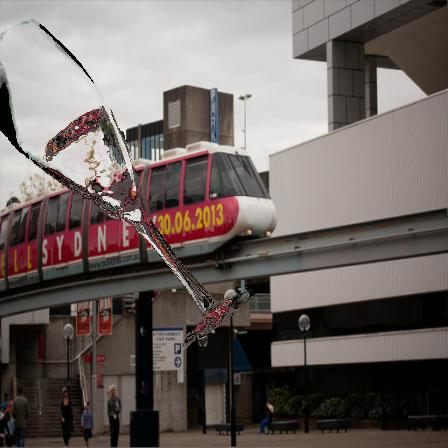}\\[-0.4 em] \midrule \\[-1.2 em]

\multirow{2}{*}[1.0 em]{\verns{Colored Lens}\hspace{0.2 cm}\addpic{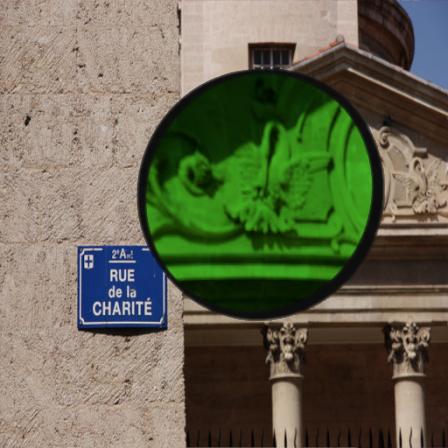}} & \multirow{2}{*}[1.0 em]{\addpic{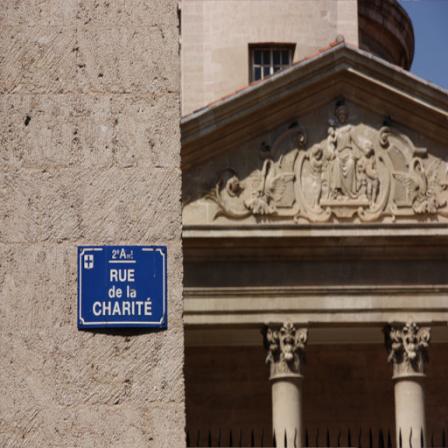}} & \ver{TOM-Net}\addpic{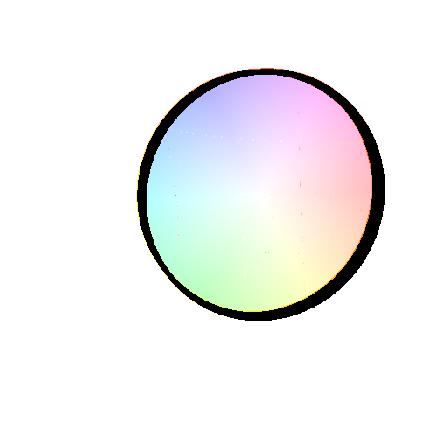} \addpic{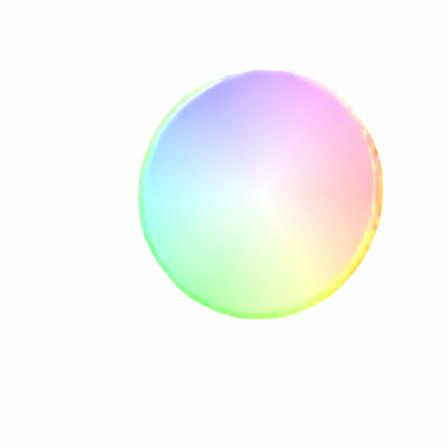} & \addpic{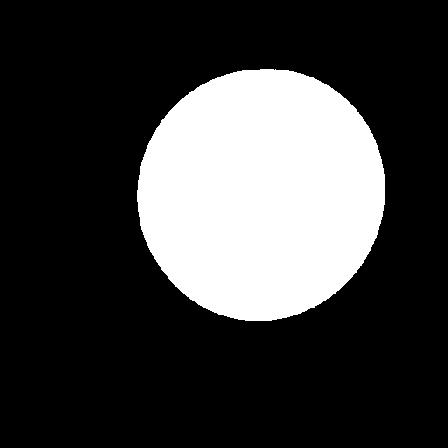} \addpic{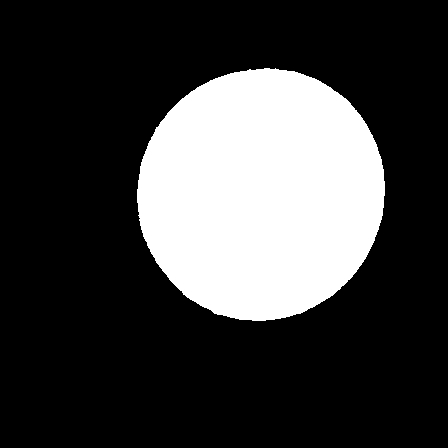} & \addpic{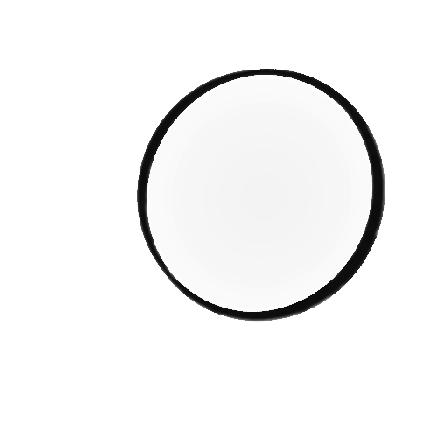} \addpic{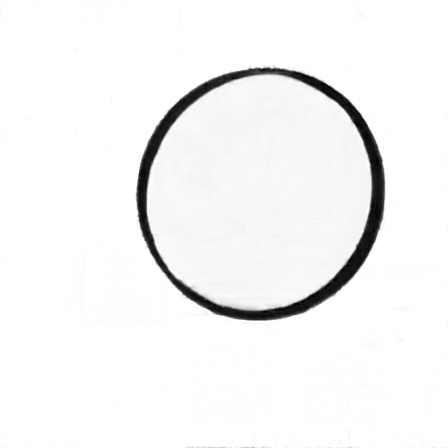} & \addpic{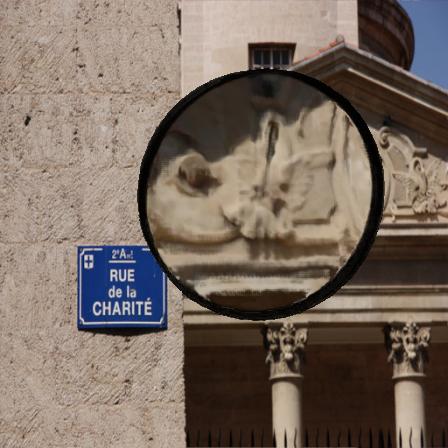} \\[-0.2 em]
&  & \ver{CTOM-Net}\addpic{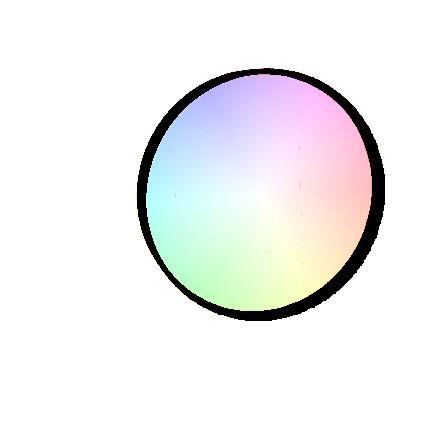} \addpic{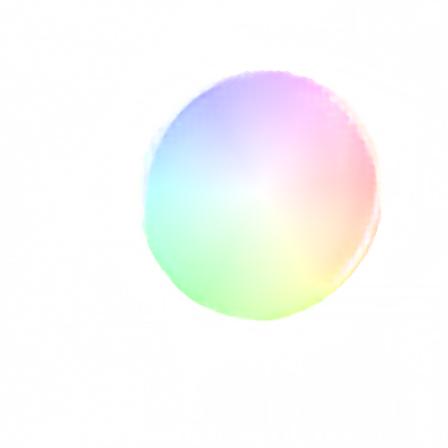} & \addpic{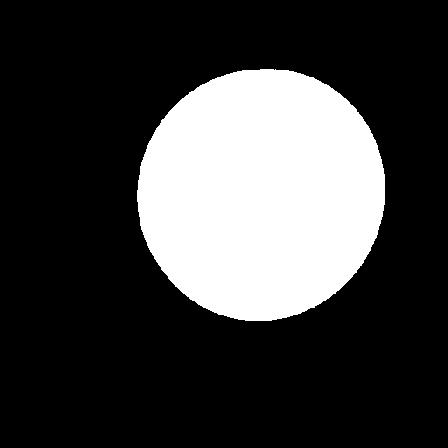} \addpic{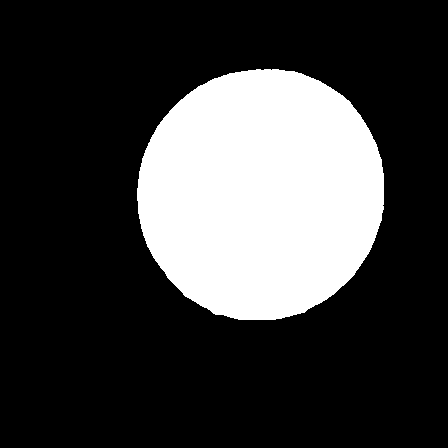} & \addpic{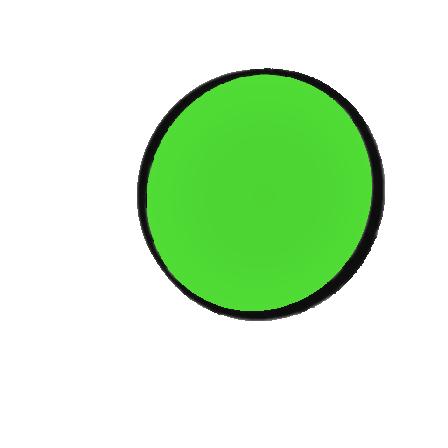} \addpic{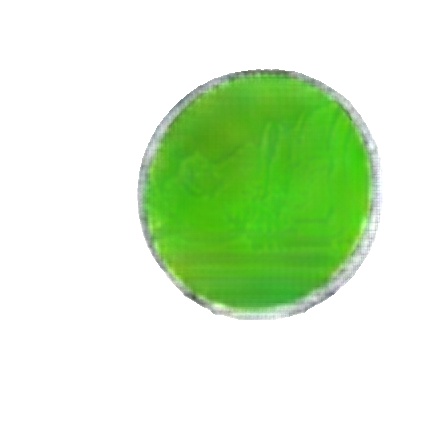} & \addpic{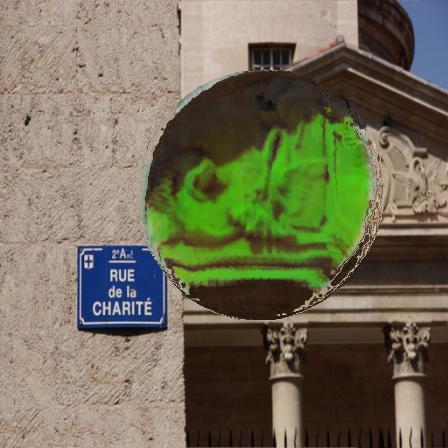}\\[-0.4 em] \midrule \\[-1.2 em]

\multirow{2}{*}[1.3 em]{\verns{Colorless Lens}\hspace{0.2 cm}\addpic{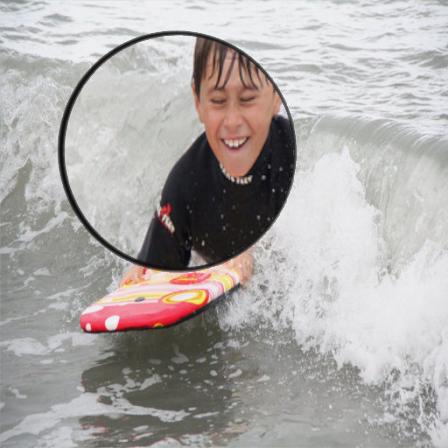}} & \multirow{2}{*}[1.0 em]{\addpic{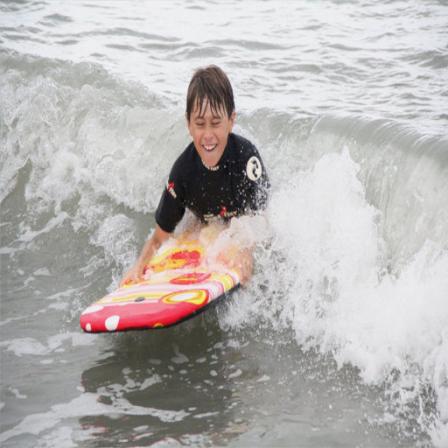}} & \ver{TOM-Net}\addpic{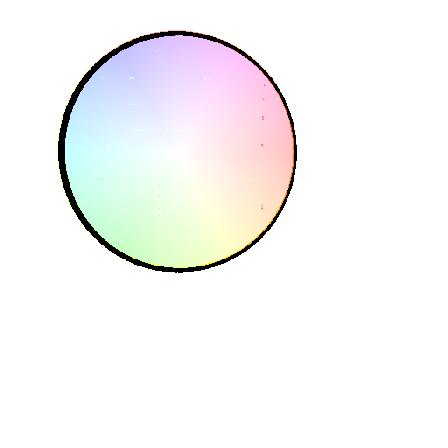} \addpic{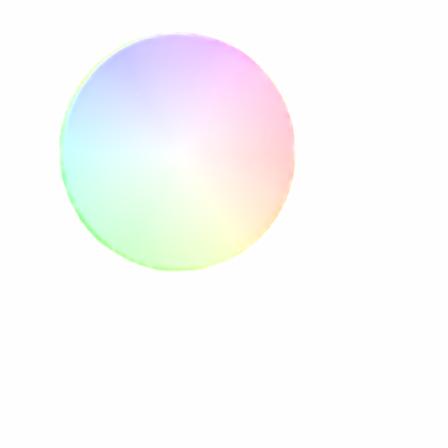} & \addpic{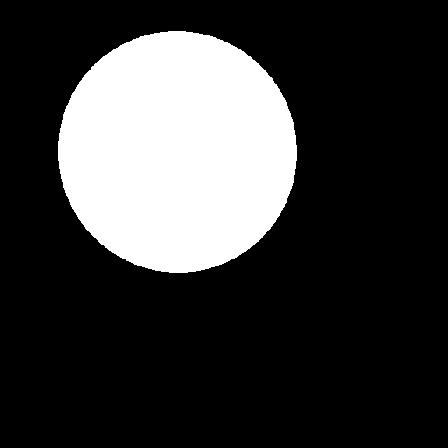} \addpic{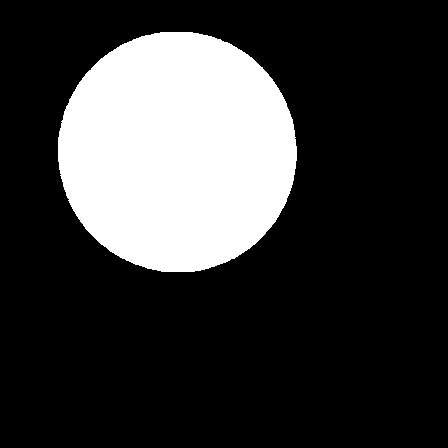} & \addpic{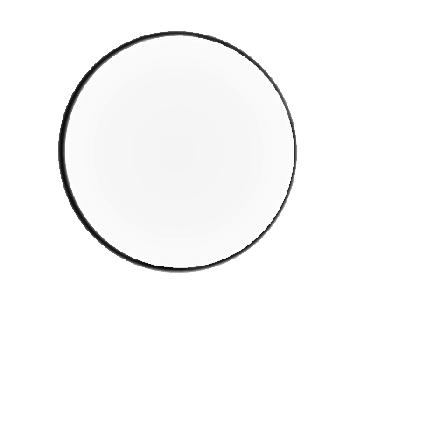} \addpic{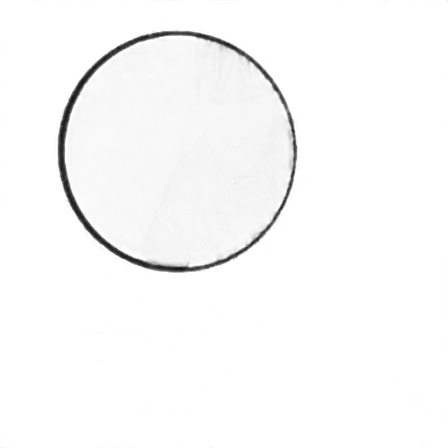} & \addpic{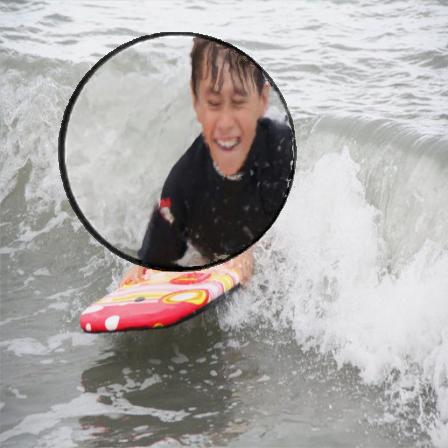} \\[-0.2 em]
&  & \ver{CTOM-Net}\addpic{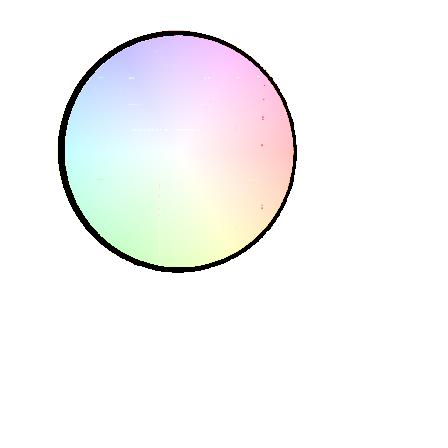} \addpic{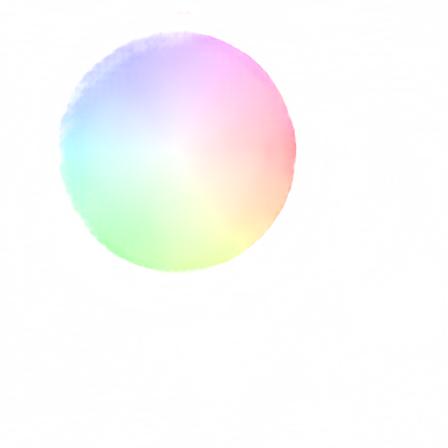} & \addpic{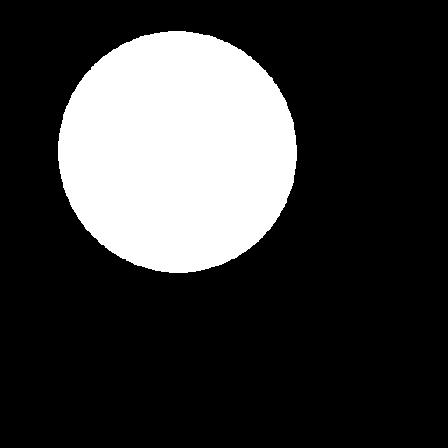} \addpic{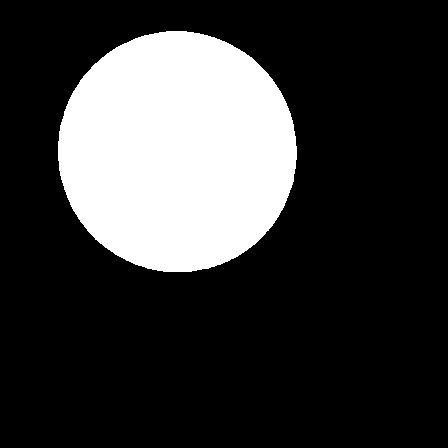} & \addpic{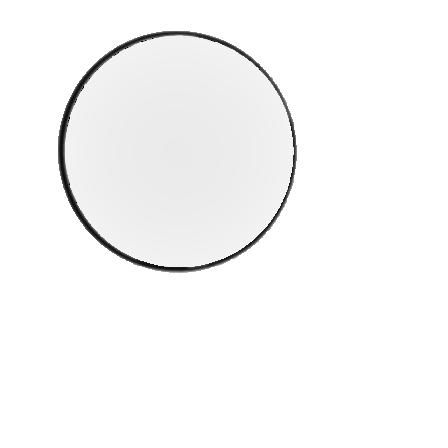} \addpic{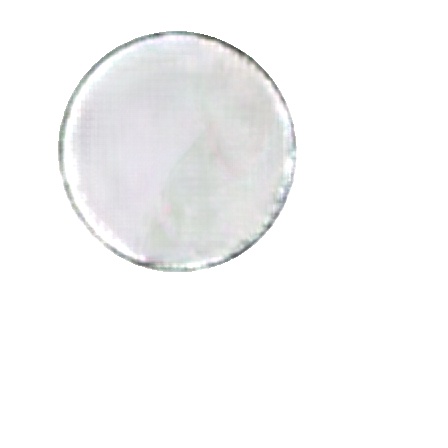} & \addpic{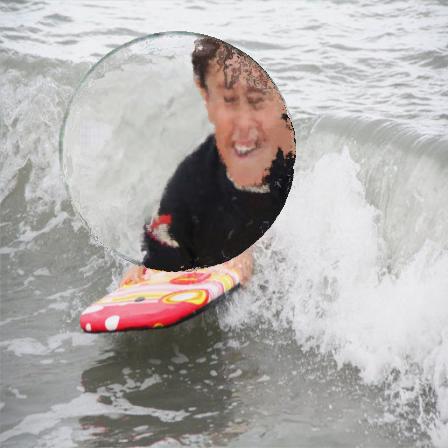}\\[-0.4 em] \bottomrule

    \end{tabularx}
        \caption{Examples of results on synthetic data, compared with TOM-Net as our baseline. For TOM-Net, we show the ground truth attenuation map and estimated attenuation map in the color filter column. GT is ground truth, and Est is the estimate made by the method. (Best viewed in PDF with zoom).}
        \label{table:synthresultfiguretable}
    \end{figure*}

For testing with the synthetic dataset, we compare our predicted mattes with the ground-truth mattes. We report the average intersection over union for the object masks, the average mean-squared error for the color filters, and the average end-point error for the refractive flow fields. We also report the average mean-squared error for the reconstructed images. We compare our results against the background images and the results of TOM-Net in Table \ref{synthetic results}. It can be observed that our model outperforms both the background images and TOM-Net. Figure \ref{table:synthresultfiguretable} shows that our model can create visually pleasing composite images on synthetic data, and can capture the color of the transparent objects well (something that cannot be handled by TOM-Net).

\begin{table*}[t!]
	\ssmall
	\centering
	\caption{Results on synthetic data. F-EPE and F-RoI are EPEs of refractive flow for whole image and region of interest respectively. M-IoU is the intersection over union for the object mask. C-MSE is the MSE of the color filter; and I-MSE is the MSE of the reconstruction. Lower is better, except for M-IoU.}
        \begin{tabularx}{\textwidth}{Y@{\hspace{1cm}}|YYYYY|YYYYY|YYYYY}
			\toprule
			\multirow{2}{*}[-0.7 em]{Method} &
			\multicolumn{5}{c|}{Colored Glass} & \multicolumn{5}{c|}{Colored Glass With Water} & \multicolumn{5}{c}{Colored Lens} \\[0.5 em]
        &F-EPE & F-RoI & M-IoU & C-MSE & I-MSE
		&F-EPE & F-RoI & M-IoU & C-MSE & I-MSE
		&F-EPE & F-RoI & M-IoU & C-MSE & I-MSE\\ \midrule
		Background & 2.55 & 28.79 & 0.09 & 4.92 & 1.17 & 4.07 & 44.57 & 0.09 & 4.88 & 1.13& 5.46 & 33.60 & 0.15 & 6.30 & 1.73\\
		\mbox{TOM-Net} &2.69 & 28.70 & 0.91 & - & 0.80&3.70 & 38.32 & 0.89 & - & 0.70 & 2.02 & 12.90 & 0.98 & - & 0.95 \\
		\mbox{CTOM-Net} & \textbf{1.55} & \textbf{17.02} & \textbf{0.98} & \textbf{0.64} & \textbf{0.26} & \textbf{2.53} & \textbf{27.94} & \textbf{0.97} & \textbf{0.75} & \textbf{0.33} & \textbf{1.05} & \textbf{6.53} & \textbf{0.99} & \textbf{1.55} & \textbf{0.67} \\
		\midrule
		\multirow{2}{*}[-0.7 em]{Method} &
		\multicolumn{5}{c|}{Colorless Glass} & \multicolumn{5}{c|}{Colorless Glass With Water} & \multicolumn{5}{c}{Colorless Lens} \\[0.5 em]
	&F-EPE & F-RoI & M-IoU & C-MSE & I-MSE
	&F-EPE & F-RoI & M-IoU & C-MSE & I-MSE
	&F-EPE & F-RoI & M-IoU & C-MSE & I-MSE\\ \midrule
	Background & 2.53 & 28.83 & 0.09 & 0.94 & 0.35 & 3.84 & 42.86 & 0.09 & 1.54 & 0.53 & 5.57 & 33.81 & 0.15 & 2.06 & 1.06\\
	\mbox{TOM-Net} & 1.90 & 20.05 & 0.91 & - & \textbf{0.20} & 3.19 & 35.05 & 0.92 & - & \textbf{0.28} & 2.03 & 13.00 & 0.99 & - & \textbf{0.35}\\
	\mbox{CTOM-Net} & \textbf{1.68} & \textbf{18.62} & \textbf{0.94} & \textbf{0.75} & 0.23 & \textbf{2.75} & \textbf{30.53} & \textbf{0.93} & \textbf{0.84} & 0.36 & \textbf{1.49} & \textbf{9.62} & \textbf{0.99} & \textbf{1.30} & 0.55\\ \bottomrule

    \end{tabularx}

        \label{synthetic results}
    \end{table*}

\begin{figure*}[t!]
	\renewcommand{\thefigure}{3}
	\centering
        \begin{tabularx}{\textwidth}{YYYYYYY}
			\toprule
        Input & Background & Ref. Flow & Mask & Color Filter & Reconstruction & Composite\\ \midrule
		% \verns{Colored Glass}\hspace{0.2 cm}
		\addpic{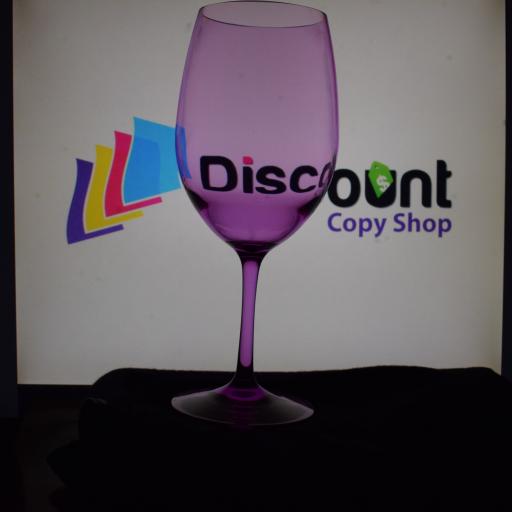} & \addpic{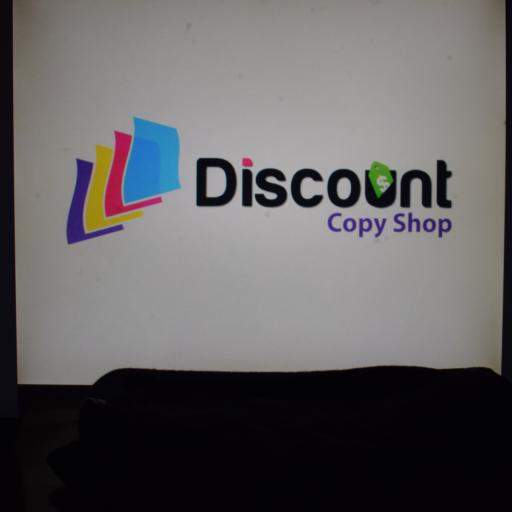} & \addpic{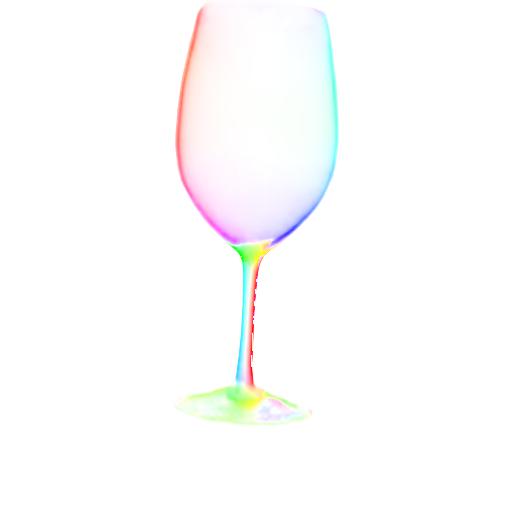} & \addpic{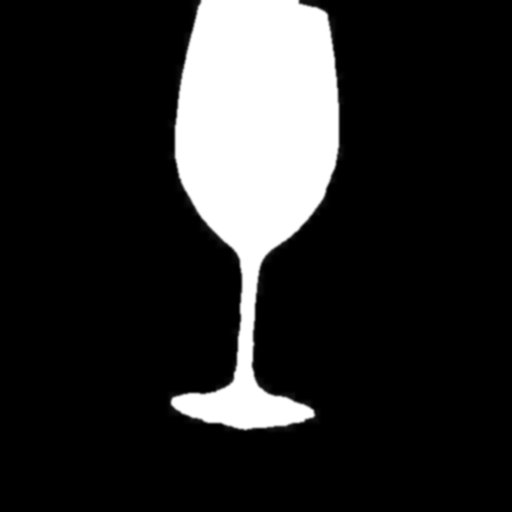} & \addpic{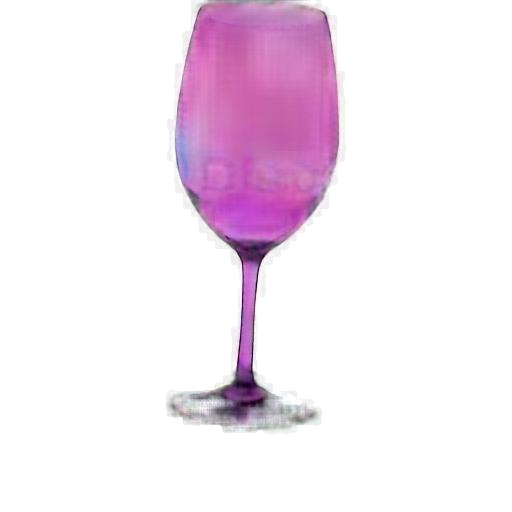} & \addpic{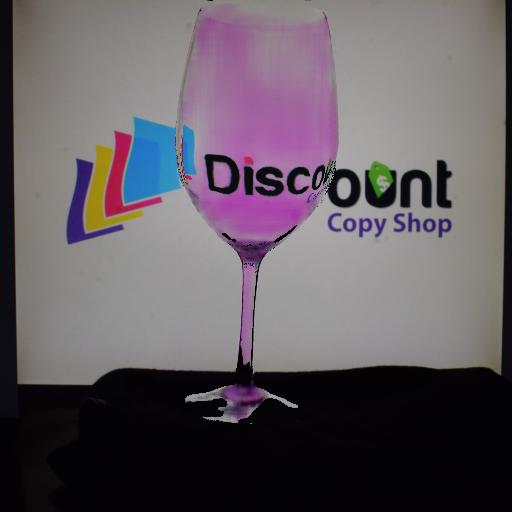} & \addpic{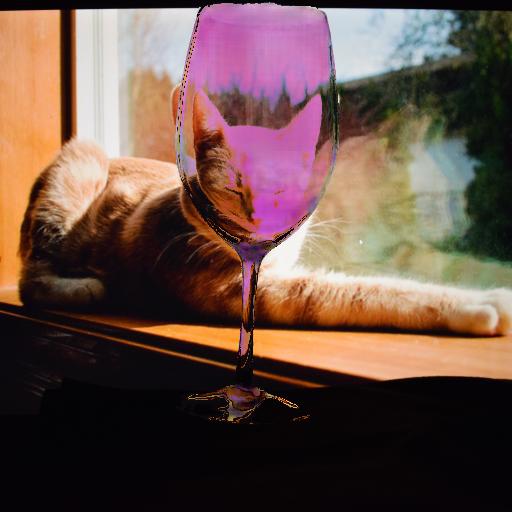} \\

		% \vernns{Colorless Glasses}\hspace{0.2 cm}
		\addpic{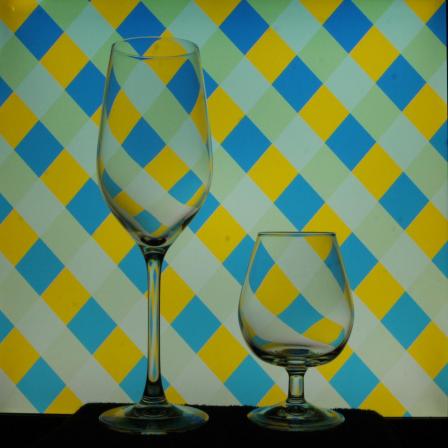} &
		% \vspace{0.3 cm}
		\addpic{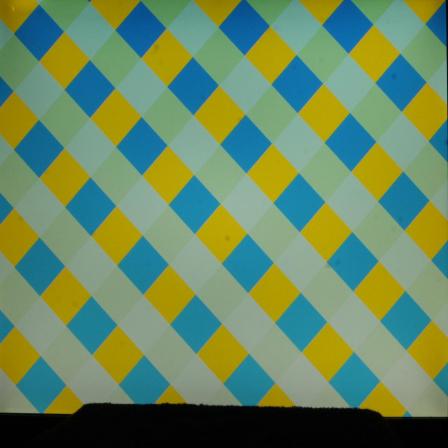} &
		% \vspace{0.3 cm}
		\addpic{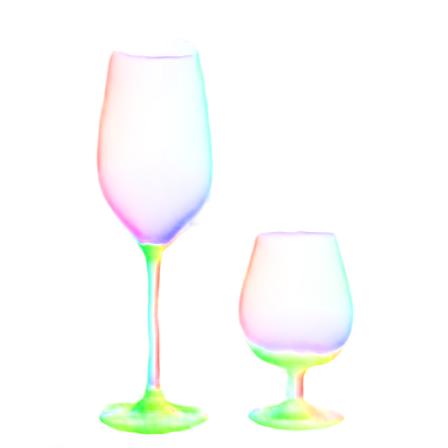} &
		% \vspace{0.3 cm}
		\addpic{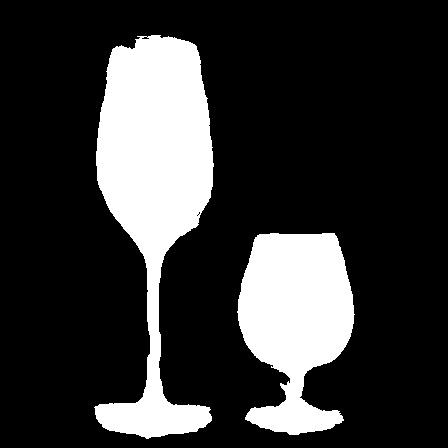} &
		% \vspace{0.3 cm}
		\addpic{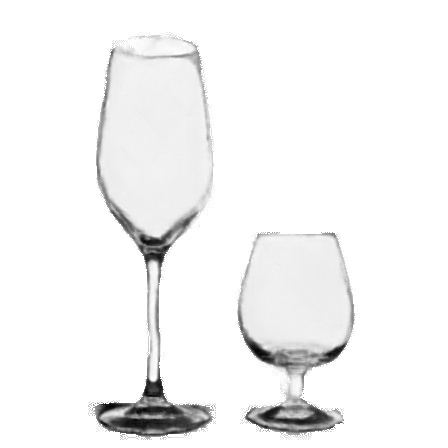} &
		% \vspace{0.3 cm}
		\addpic{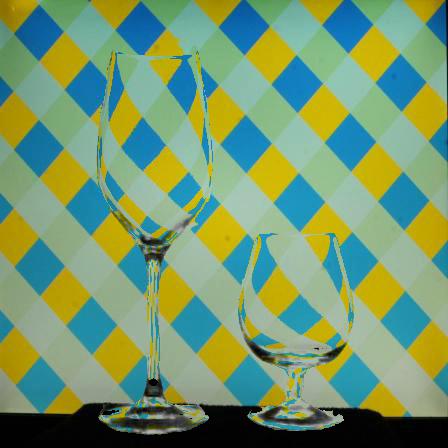} &
		% \vspace{0.3 cm}
		\addpic{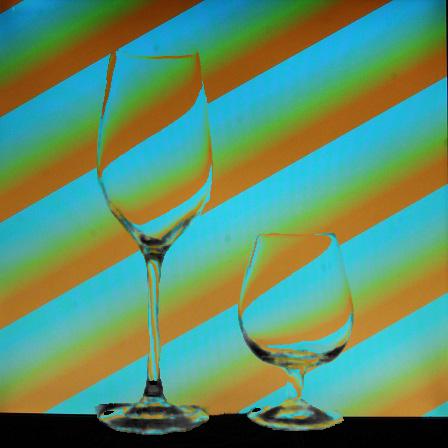} \\

		\addpic{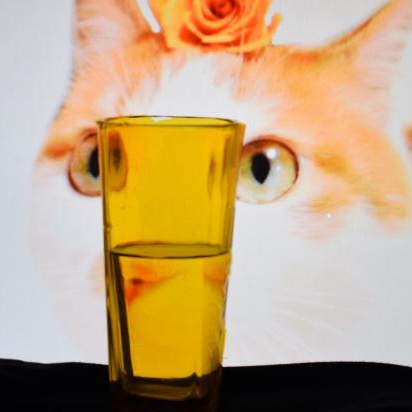} & \addpic{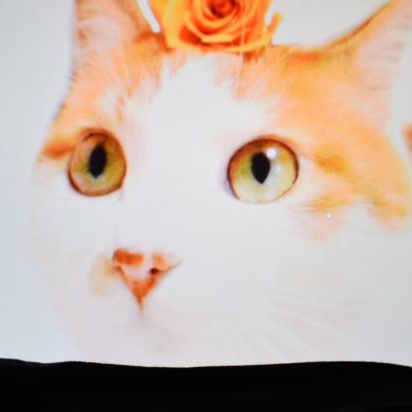} & \addpic{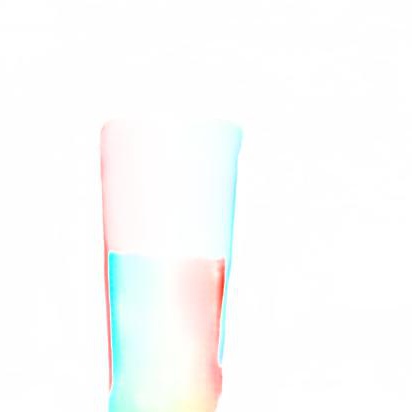} & \addpic{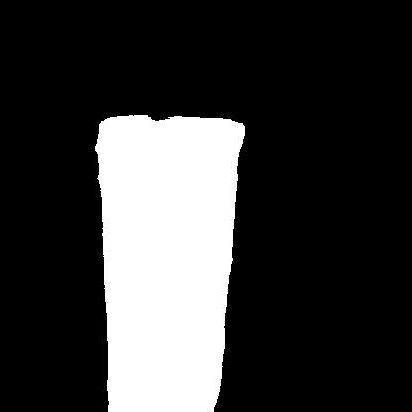} & \addpic{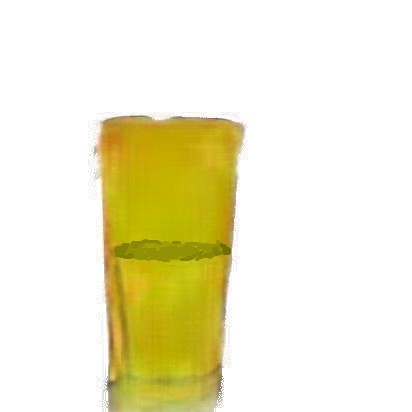} & \addpic{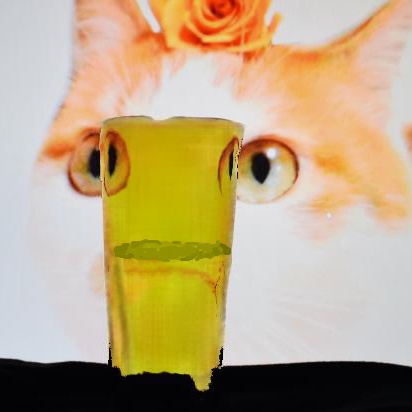} &
		\addpic{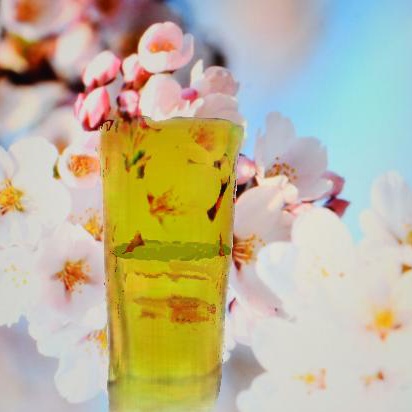}
		\\

		% \verns{Colored Lens}\hspace{0.2 cm}
		\addpic{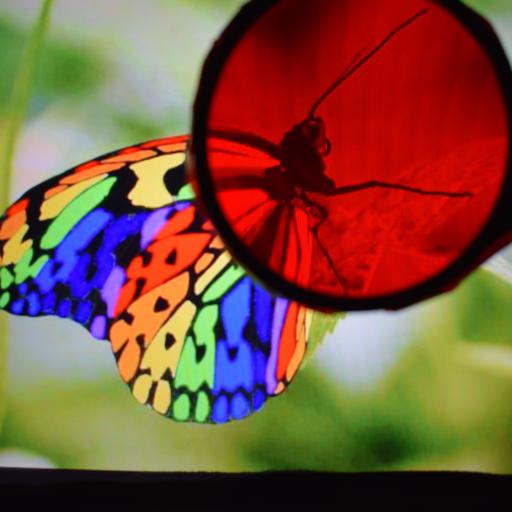} & \addpic{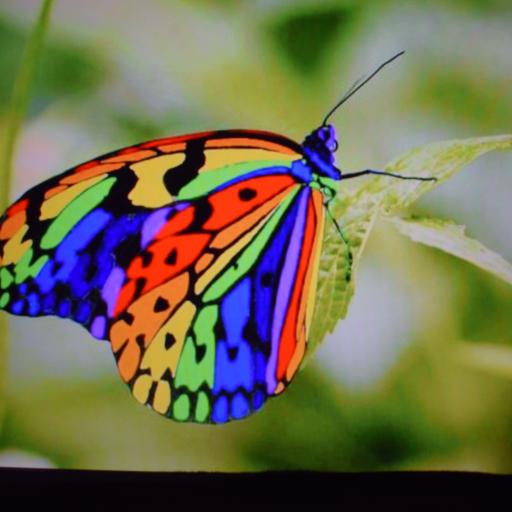} & \addpic{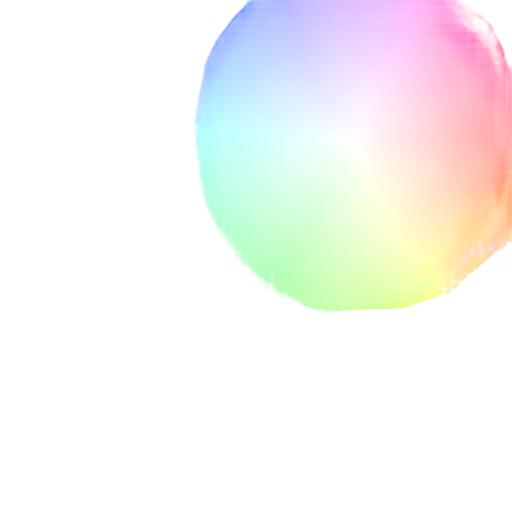} & \addpic{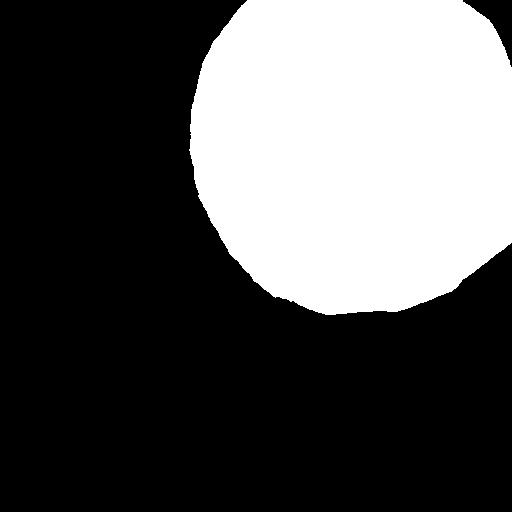} & \addpic{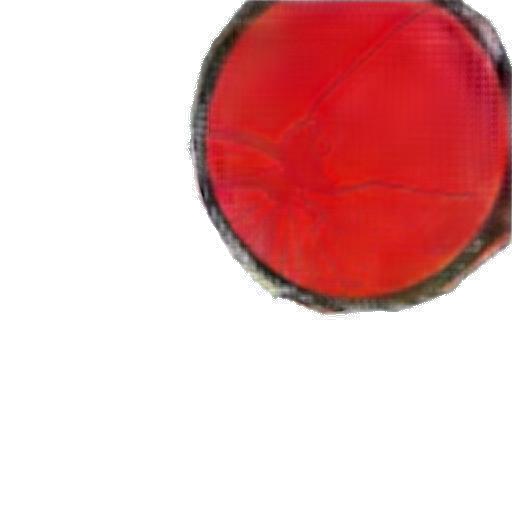} & \addpic{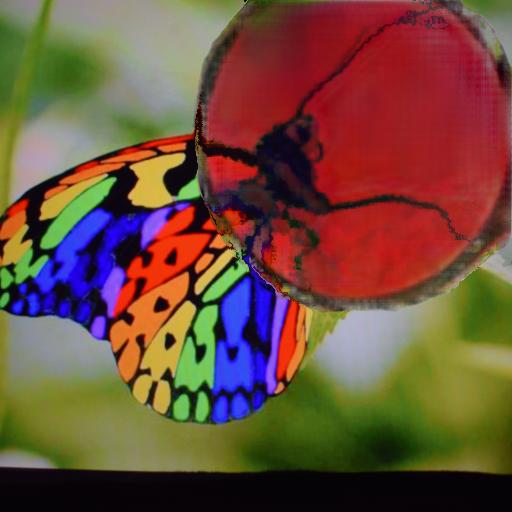} & \addpic{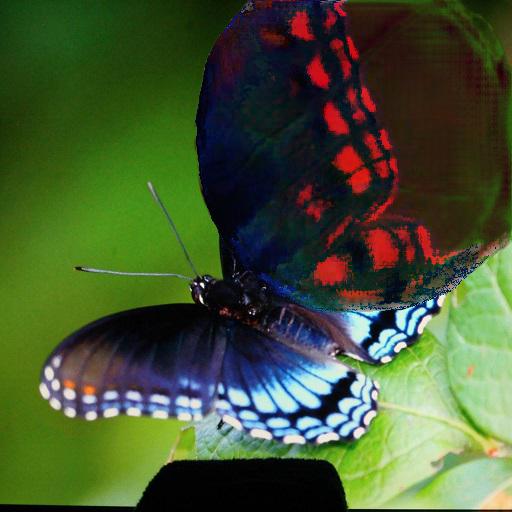} \\

		\bottomrule
    \end{tabularx}
        \caption{Examples of results on real data. Composites are made by rendering the extracted object against a new background.}
        \label{table:realresultfiguretable}
    \end{figure*}

\subsection{Comparison with TOM-Net}
In comparison to TOM-Net, our model makes more robust estimates of the refractive flow field on both colorless and colored transparent objects. Table \ref{synthetic results} shows that our method outperforms TOM-Net in refractive flow fields estimation in all three object categories. The example of colorless glass with water in Figure \ref{table:synthresultfiguretable} shows that TOM-Net completely fails to detect the refractive flow for the water inside the glass, whereas CTOM-Net can detect the water and estimates the refractive flow. The example of colored glass with water shows that TOM-Net detects part of the building as a transparent object. This is also reflected by the average intersection over union values for the object mask across all three object categories in Table \ref{synthetic results} , where CTOM-Net achieves higher values than TOM-Net. Note that TOM-Net models only colorless transparent objects and predicts an attenuation mask, whereas CTOM-Net considers colored transparent objects and predicts a color filter. This introduces a small margin for error in image reconstruction, and the reconstructed image MSE for TOM-Net for colorless objects is therefore marginally better than that of CTOM-Net. Both methods share the same objective of producing visually pleasing results, and Figure \ref{table:synthresultfiguretable} shows no discernible difference in the image reconstruction of colorless objects. In summary, CTOM-Net is a more robust and useful (in the sense that it can handle colored transparent objects) method for transparent object matting.

\subsection{Results on Real Data} We test our trained model on the real dataset and compare the reconstructed images with the background images. Due to the absence of ground-truth values, we evaluate the results using the peak signal-to-noise ratio (PSNR) and structural similarity (SSIM) metrics. The average values shown in Table \ref{real results} show that our method outperforms background images in terms of PSNR and SSIM metrics. Since most of our testing is done on colored objects, and we know TOM-Net cannot handle such objects, and for brevity, we do not compare with TOM-Net in this section, but instead focus on showing how well CTOM-Net can generalize on real images. Figure \ref{table:realresultfiguretable} shows that for real images, our model predicts mattes that allow rendering visually pleasing and realistic images. This shows that our model can generalize well on real images.

\begin{table}[htb]
	\ssmall
	\centering
	\caption{Results for real dataset.}
        \begin{tabular}{c|cc|cc|cc}
			\toprule
			\multirow{2}{*}[-0.7 em]{Method} &
			\multicolumn{2}{c|}{Glass} &
			 \multicolumn{2}{c|}{G\&W} &
			 \multicolumn{2}{c}{Lens} \\
        &PSNR & SSIM&PSNR & SSIM&PSNR & SSIM\\ \midrule
		Background & 16.15 & 0.88 & 15.97 & 0.86 & 11.85 & 0.68\\
		CTOM-Net & 22.63 & 0.90 & 23.21 & 0.91 & 16.36 & 0.75\\
		\bottomrule

    \end{tabular}

        \label{real results}
    \end{table}

\clearpage

\section{Conclusion}
We have introduced a simple yet efficient model for colored transparent object matting, and proposed a two-stage deep learning framework to tackle this problem. Our method takes a single image as input, and predicts an environment matte as an object mask, a color filter, and a refractive flow field for a transparent object. Experiments on both synthetic and real datasets show that our proposed method achieves promising results, clearly demonstrating the effectiveness of our method.

Since our method assumes the background image is the only light source, our method cannot handle transparent objects with specular highlights. In the future, we will investigate better training datasets and network architectures to extend our framework to handle objects with specular highlights.
%We proposed a matte for transparent objects with or without intrinsic color and train a neural network ensemble to predict such mattes. Our method fulfils our aim of producing visually pleasing results and works well on real images.

%Future work can be done in order to make the model generalize well on more complex transparent objects, and to model any specular reflection on transparent objects.

%\section{Acknowledgments:}
%  Omitted for anonymity.

\end{document}